\documentclass[acmsmall]{acmart}

\usepackage{amsmath,amsfonts}
\usepackage{algorithm,algorithmic}
\usepackage{color}
\usepackage{balance}
\usepackage{hyperref}

\usepackage{amsmath,amsfonts,amssymb,multirow}
\usepackage{graphicx, subfigure, epstopdf, epsfig}

\AtBeginDocument{%
  }

\setcopyright{acmcopyright}
\copyrightyear{2018}
\acmYear{2018}
\acmDOI{XXXXXXX.XXXXXXX}

\acmJournal{JACM}
\acmVolume{37}
\acmNumber{4}
\acmArticle{111}
\acmMonth{8}

\begin{document}

\title{Binary Representation via Jointly Personalized Sparse Hashing}

\author{Xiaoqin Wang}
\email{xqwang@guet.edu.cn}
\author{Chen Chen}
\email{chenchen_guet@163.com}
\author{Rushi Lan}
\email{rslan@guet.edu.cn}
\affiliation{%
  \institution{{Guangxi Key Laboratory of Image and Graphic Intelligent Processing, Guilin University of Electronic Technology}
  \city{Guilin}
  \country{China}
}}

\author{Licheng Liu}
\affiliation{%
  \institution{College of Electrical and Information Engineering, Hunan University}
  \city{Changsha}
  \country{China}
}
\email{lichenghnu@gmail.com}

\author{Zhenbing Liu}
\affiliation{%
  \institution{Guangxi Key Laboratory of Image and Graphic Intelligent Processing, Guilin University of Electronic Technology}
  \city{Guilin}
  \country{China}
}
\email{zbliu2011@163.com}

\author{Huiyu Zhou}
\affiliation{%
  \institution{School of Computing and Mathematical Sciences, University of Leicester}
  \city{Leicester}
  \country{the United Kingdom}
}
\email{hz143@leicester.ac.uk}

\author{Xiaonan Luo}
\affiliation{%
  \institution{Guangxi Key Laboratory of Image and Graphic Intelligent Processing, Guilin University of Electronic Technology}
  \city{Guilin}
  \country{China}
}
\email{luoxn@guet.edu.cn}

\thanks{Corresponding authors: Rushi Lan and Licheng Liu}

\renewcommand{\shortauthors}{Xiaoqin Wang et al.}

\begin{abstract}
Unsupervised hashing has attracted much attention for binary representation learning due to the requirement of economical storage and efficiency of binary codes. It aims to encode high-dimensional features in the Hamming space with similarity preservation between instances. However, most existing methods learn hash functions in manifold-based approaches. Those methods capture the local geometric structures (i.e., pairwise relationships) of data, and lack satisfactory performance in dealing with real-world scenarios that produce similar features (e.g. color and shape) with different semantic information. To address this challenge, in this work, we propose an effective unsupervised method, namely Jointly Personalized Sparse Hashing (JPSH), for binary representation learning. To be specific, firstly, we propose a novel personalized hashing module, i.e., Personalized Sparse Hashing (PSH). Different personalized subspaces are constructed to reflect category-specific attributes for different clusters, adaptively mapping instances within the same cluster to the same Hamming space. In addition, we deploy sparse constraints for different personalized subspaces to select important features. We also collect the strengths of the other clusters to build the PSH module with avoiding over-fitting. Then, to simultaneously preserve semantic and pairwise similarities in our proposed JPSH, we incorporate the proposed PSH and manifold-based hash learning into the seamless formulation. As such, JPSH not only distinguishes the instances from different clusters, but also preserves local neighborhood structures within the cluster. Finally, an alternating optimization algorithm is adopted to iteratively capture analytical solutions of the JPSH model. We apply the proposed representation learning algorithm JPSH to the similarity search task. Extensive experiments on four benchmark datasets verify that the proposed JPSH outperforms several state-of-the-art unsupervised hashing algorithms.
\end{abstract}

%



\maketitle

\section{Introduction}
Representation learning has attracted extensive research attention in affective applications, such as affective image retrieval \cite{yao2019attention,yao2020adaptive,zhao2014affective}, emotion classification \cite{chen2015learning,rao2020learning,rao2019multi} and facial recognition \cite{shi2020towards}. Traditional representation learning usually bases on real-valued features, which will cause a waste of time and space. Recently, hashing representation also known as binary representation has been popular and achieved promising performances \cite{xiang2021sub, jin2018deep, tang2018discriminative}. Furthermore, unsupervised hashing refers to the technology that maps high-dimensional features to compact binary codes without label information \cite{lib2021structure,liu2014discrete,hao2017unsupervised,hu2020creating,zhang2020deep}, and the similarity relationship between instances can be approximated by the Hamming distance. Therefore, unsupervised hashing with the low storage and efficient computation can be widely used in representation learning tasks, especially in the image similarity search tasks, which aims to retrieve some related samples from the dataset \cite{zhu2018exploring,jin2021unsupervised}. 

Existing unsupervised hashing methods can be categorized into data-independent and learning-based hashing. Previous works mainly focus on finding suitable projections to produce optimal binary codes, e.g., Locality Sensitive Hashing (LSH) \cite{datar2004locality} and Min-wise Hashing (Min-Hash) \cite{indyk2001small}. Such methods model learning processes without using any data structure and distribution in the original space, and require long hash bits to achieve satisfactory results.

In contrast, learning-based hashing methods are gaining popularity in recent years. These methods preserve similarity relationships between instances via different viewpoints. For example, Iterative Quantization with Principal Component Analysis (PCA) (PCA-ITQ) \cite{gong2012iterative} imposes a rotation matrix to iteratively reduce quantization errors between low-dimensional features and binary codes. Sparse Projections (SP) \cite{xia2015sparse} incorporates a sparse regularizer to reduce the number of parameters and computational cost. Ordinal Embedding Hashing (OEH) \cite{liu2016towards} and Ordinal Constraint Hashing (OCH) \cite{liu2018ordinal} preserve the ranking information by embedding the ordinal relation among data points. Concatenation Hashing (CH) \cite{weng2020concatenation} encourages that any two instances close to the same center point are close, whilst maintaining relative positions in the Hamming space. Recovery of Subspace Structures Hashing (RSSH) \cite{tian2020unsupervised} preserves the semantic similarity in the Hamming space via an unsupervised multi-stage hashing model. 

In addition, some manifold-based hashing methods have been proposed to capture the complex structures of data based on graph learning. Spectral Hashing (SH) \cite{weiss2008spectral} converted the binary code learning to the graph partitioning. Zhu et al. proposed the Sparse Embedding and Least Variance Encoding (SELVE) \cite{zhu2014sparse} to encode the sparse embedding vector over a learned dictionary. Anchor Graph Hashing (AGH) \cite{liu2011hashing} built an anchor graph \cite{liu2010large}, and it obtained the tractable low-rank adjacency matrix. The matrix was used to measure the similarity between a pair of data points and a small number of anchor points, which were the K-means clustering centers because of the strong representation power of the category attributes. To preserve the underlying manifold structure with the t-SNE \cite{van2008visualizing} that is a modification of stochastic neighborhood embedding, Inductive Manifold Hashing (IMH) \cite{shen2013inductive} was propsed. Some researchers developed a tractable alternating maximization algorithm to preserve the neighborhood structure inherent in the data, called Discrete Graph Hashing (DGH) \cite{liu2014discrete}. Jiang et al. implicitly computed the similarity graph matrix by feature transformation, and then proposed the Scalable Graph Hashing (SGH) \cite{jiang2015scalable} method. Graph PCA (gPCA) Hashing \cite{zhu2017graph} simultaneously preserved local structures via manifold learning and global structures via PCA. Locally Linear Hashing (LLH) \cite{irie2014locally} reconstructed the locally linear structures of manifolds in the binary Hamming space with locality-sensitive coding. Compared to LLH, Discrete Locality Linear embedding Hashing (DLLH) \cite{ji2017toward} directly reconstructed binary codes by maintaining the local linear relationship of data points. Considering the $l_{2,1}$-norm term, Jointly Sparse Hashing (JSH) \cite{lai2018jointly} could minimize the information loss. Unsupervised Discrete Hashing (UDH) \cite{jin2021unsupervised} captured the semantic information by a balanced graph semantic loss for exploring both the similar and dissimilar relationships among data. By considering the semantic information, Li et al. proposed a weakly-supervised hashing method for image retrieval and achieved state-of-the-art performance \cite{li2020weakly}.

\begin{figure}[h]
	\centering
	\includegraphics[width=\linewidth]{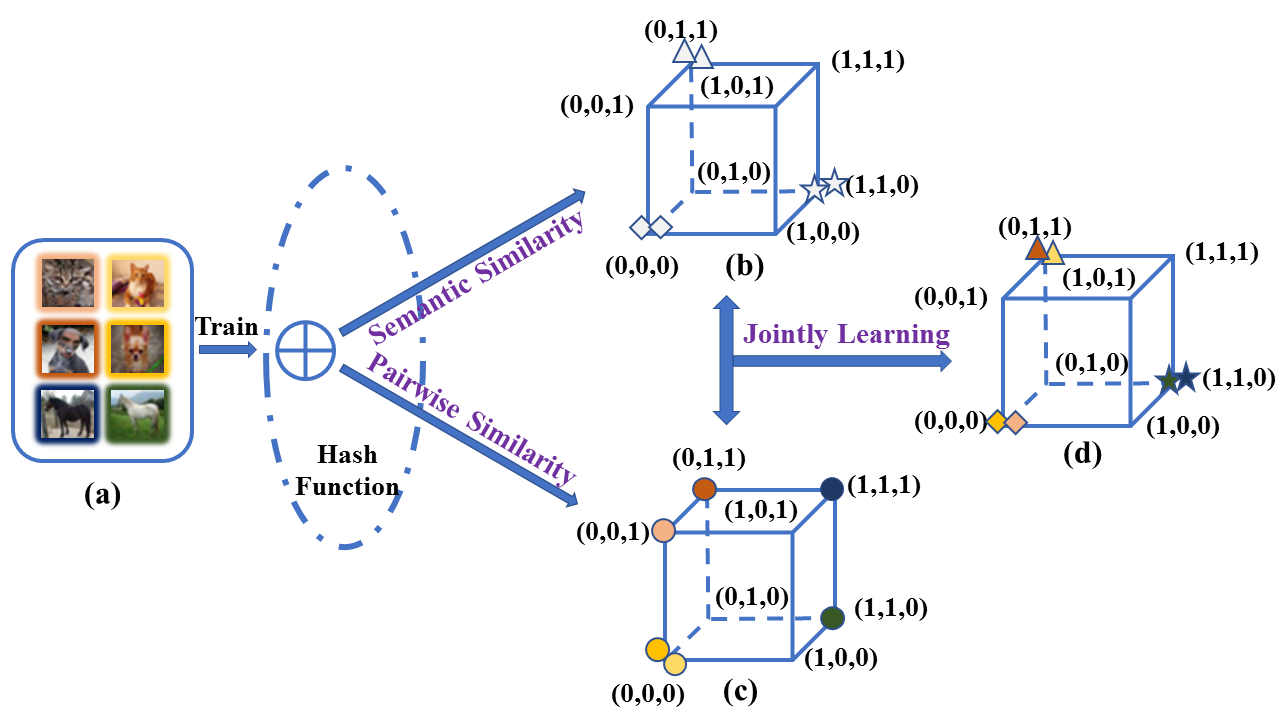}
	\caption{The proposed JPSH framework. It constructs a seamless hash function, which consists of twofold properties: semantic and pairwise similarities. JPSH accommodates the proposed PSH module to maintain semantic similarity, and preserves pairwise similarity using a manifold-based hashing method. Thus, we learn discriminative binary codes by combining the two similarities. In this figure, shapes including triangle, square and star are used to describe semantic relationships except for circle, whilst colors are used to describe pairwise relationships. (a) Some real-world instances with similar attributes have different semantic information. (b) Preserving semantic similarity by the proposed PSH, adaptively mapping instances within the same cluster to the same Hamming space. (c) Maintaining the pairwise similarity by a manifold-based hashing method. (d) Jointly learning the above  twofold properties in the same Hamming space, and obtaining the discriminative hashing codes for the image binary representation.}
	\label{fig:1}
\end{figure}

However, most of existing manifold-based hashing methods do not always produce satisfactory results in practice because instances with similar features (e.g. color and shape) may have different semantics. An example is shown in Figure \ref{fig:1} (a), where a cat (row 1, column 2) is visually similar to a dog (row 2, column 2) instead of the other cat (row 1, column 1). On the other hand, there is a large difference in color between the black horse (row 3, column 1) and the white horse (row 3, column 2), which may be mistakenly treated as two different animals. Existing methods model the neighboring graph in hash functions to reflect the pairwise relationship between two instances. For example, SH constructs the Laplacian graph to describe the relationship between two instances. AGH creates a modified version of SH by introducing the anchor graph \cite{liu2010large}. JSH uses a $l_{2,1}$-regularized term to minimize the information loss of low-dimensional features and binary codes on the basis of AGH. Although the above methods can effectively preserve the local neighborhood structure from the high-dimensional feature space (shown in Figure 2 (a)) to a low-dimensional space (shown in Figure 2 (b)), such methods fail to consider semantics of instances that correspond to the intrinsic geometric structures of data \cite{zhang2019sadih,yang2018semantic,lin2015semantics,fernandez2020unsupervised}.
\begin{figure}[h]
	\centering
	\includegraphics[width=\linewidth]{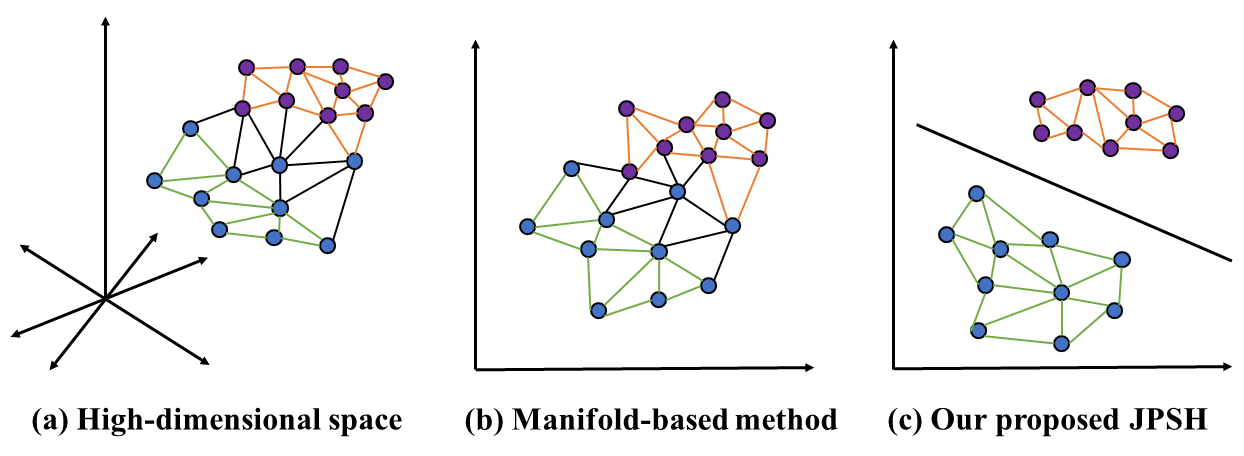}
	\caption{Comparison of the standard manifold-based method \cite{lai2018jointly} and our JPSH method. (a) the data distribution in the high-dimensional feature space; (b) the low-dimensional features generated by a standard manifold-based hashing method \cite{lai2018jointly}; (c) the low-dimensional features with discriminative achieved by our proposed JPSH, and each cluster is transformed by a personalized weight produced by the PSH module. Different from the standard scheme \cite{lai2018jointly} that only preserves pairwise similarity, the proposed JPSH jointly preserves semantic and pairwise similarities in the low-dimensional space, which allows us to effectively distinguish different clusters, and preserve local intrinsic structures within each cluster.}
	\label{fig:2}
\end{figure}

To tackle the above problems, in this paper, we propose an effective unsupervised hashing method, namely Jointly Personalized Sparse Hashing (JPSH), to learn the binary representation for similarity search. Unlike existing manifold hashing methods that only preserve the pairwise relationship, the proposed JPSH encodes both semantic and pairwise similarities into the Hamming space, which can effectively distinguish different clusters while preserving local structures within clusters, as illustrated in Figure 2 (c). The proposed JPSH framework is shown in Figure \ref{fig:1}, and the main contributions of our work can be summarized as follows:
\begin{itemize}
	\item A novel Personalized Sparse Hashing (PSH) module is proposed to preserve the semantics of instances that correspond to the intrinsic structures of data. It learns personalized subspaces that reflect the special categorical characteristics, and then defines sparse constraints for each personalized subspaces. Thus, the PSH can adaptively map instances within the same cluster to the same Hamming space and learn more discriminative features. To our knowledge, this is the first work that preserves semantic similarity in a personalized subspace.
	\item We incorporate PSH and a standard manifold-based hashing method JSH in a seamless formulation to construct the proposed JPSH model, which simultaneously preserves semantic and pairwise similarities in the Hamming space.	
\end{itemize}

The rest of this paper is organized as follows. Section~\ref{Related Work} introduces some related works. Section~\ref{Jointly Personalized Sparse Hashing} presents the details of our JPSH including the proposed personalized sparse hashing, the overall loss function, the discussions, the optimization algorithm, the out-of-sample extension. Section~\ref{Experimental Work} shows experimental results and analyses, followed by the conclusion in Section~\ref{Conclusion}.

\section{Related Works}
\label{Related Work}
In this section, we first define some used notations, and then introduce several related works, including Spectral Hashing, Anchor Graph Hashing and Jointly Sparse Hashing.

\subsection{Notations}
In the following, we use bold uppercase letters (e.g. $\boldsymbol{\rm Z}$) for matrices, bold lowercase letters (e.g. $\boldsymbol{\rm z}$) for vectors, and normal lowercase letters (e.g. z) for scalars. If there are no special instructions, we define the $i$-th row of the matrix $\boldsymbol{\rm Z}$ as $\boldsymbol{\rm Z}_{i*}$, the $j$-th column of the matrix $\boldsymbol{\rm Z}$ as $\boldsymbol{\rm Z}_{*j}$, the $i$-th element of $\boldsymbol{\rm Z}_{*j}$ or the $j$-th element of $\boldsymbol{\rm Z}_{i*}$ as $\boldsymbol{\rm Z}_{ij}$. We also denote the transpose of the matrix $\boldsymbol{\rm Z}$ as $\boldsymbol{\rm Z}^T$, and the trace of the matrix $\boldsymbol{\rm Z}$ as ${\rm tr}\left(\boldsymbol{\rm Z}\right)$ if it is a square matrix. $\|\boldsymbol{\rm Z}\|_{2}$, $\|\boldsymbol{\rm Z}\|_{2,1}$, and $\|\boldsymbol{\rm Z}\|_{F}$ represent the $l_{2}$-norm, $l_{2,1}$-norm, and Frobenius norm of the matrix $\boldsymbol{\rm Z}$, respectively. $\boldsymbol {\rm Z} \otimes \boldsymbol {\rm H}$ denotes the Kronecker product between matrices $\boldsymbol{\rm Z}$ and $\boldsymbol{\rm H}$. In this paper, the training set is denoted as $\boldsymbol{\rm X} = \{\boldsymbol{\rm x}_i\}_{i=1}^n$, $\boldsymbol{\rm x}_i\in {\rm R}^d$, where $n$ is the number of training samples, and $d$ is the dimension of features. The binary codes $\boldsymbol{\rm B} = \{\boldsymbol{\rm b}_i\}_{i=1}^m$, $\boldsymbol{\rm b}_i\in {\rm R}^l$, where $l\ll d$, $l$ is the length of bits, and $m\ll n$, $m$ is the number of anchor points. $\boldsymbol{\rm I}$ is the identity matrix with the size of $l$-by-$l$. 

\subsection{Spectral Hashing}
SH \cite{weiss2008spectral} aims to extract the discriminative information from the original space to learn binary codes based on the graph theory. With this objective, it minimizes the errors of binary codes between instances with Laplacian graph as follows:
\begin{equation}\label{eq1}
	\begin{aligned}
		\mathop{\min}_{\substack{\boldsymbol{\rm H}}}
		&\sum_{i,j=1}^{n}\|\boldsymbol{\rm h}_i-\boldsymbol{\rm h}_j\|_2^2\boldsymbol{\rm M}_{ij}\\
		&s.t.\  \boldsymbol{\rm H}\in \{-1,1\}^{l\times n}, \boldsymbol{\rm l}\boldsymbol{\rm H} = \boldsymbol{\rm 0}, \boldsymbol{\rm H}\boldsymbol{\rm H}^T = n\boldsymbol{\rm I}\\
	\end{aligned},
\end{equation}
where $\boldsymbol{\rm l}\in R^{1\times l}$ is an all-ones vector, $\boldsymbol{\rm M}_{ij}={\rm exp}\left(-\|\boldsymbol{\rm x}_i-\boldsymbol{\rm x}_j\|^2/\eta^2\right)$, $\boldsymbol{\rm M}\in R^{n\times n}$ is the affinity matrix, $\eta$ is a pre-defined parameter, and $\boldsymbol{\rm H}=\{\boldsymbol{\rm h}_i\}_{i=1}^n, \boldsymbol{\rm h}_i\in {\rm R}^l$ is binary codes of the $i$-th training sample. It is time-consuming to compute $\boldsymbol{\rm M}$ for large-scale datasets. In addition, the information loss between low-dimensional features and binary codes will increase due to the binarization operation.

\subsection{Anchor Graph Hashing}
To reduce the computational complexity in Eq.~\eqref{eq1}, AGH \cite{liu2011hashing} designs a truncated similarity matrix $\boldsymbol{\rm A}\in {\rm R}^{n\times m}$ between all $n$ instances and $m$ anchor points. The $i$-th element of $\boldsymbol{\rm A}_{*j}$ is defined as:
\begin{equation}\label{eq2}
	\boldsymbol{\rm A}_{ij}=\left\{
	\begin{aligned}
		&\frac{{\rm exp}(-\|\boldsymbol{\rm x}_i-\boldsymbol{\rm c}_j\|^2/\theta)}{\sum_{j'\in \{i\}}{\rm exp}(-\|\boldsymbol{\rm x}_i-\boldsymbol{\rm c}_{j'}\|^2/\theta)} \  ,\forall j\in \{i\} \\
		&0 \quad \quad \quad \quad \quad \quad \quad \quad \quad \quad \quad \quad \ \ {\rm ,otherwise}
	\end{aligned},
	\right.
\end{equation}
where $\boldsymbol{\rm c}_j\in {\rm R}^d$ is the $j$-th anchor point; $\{i\}$ is the index set of $k$ neighboring anchor points of $\boldsymbol{\rm x}_i$, and $\theta$ is a pre-defined parameter. Although AGH has the linear training time for the developed model, it still fails to control the information loss between low-dimensional features and binary codes in the binary representation learning.

\subsection{Jointly Sparse Hashing}
To minimize the information loss between low-dimensional feature vectors and binary codes, JSH \cite{lai2018jointly} presents $l_{2,1}$-regularized regression formulation, and the objective function can be defined as follows:
\begin{equation}\label{eq3}
	\begin{aligned}
		\mathop{\min}_{\substack{\boldsymbol{\rm B},\boldsymbol{\rm W},\boldsymbol{\rm V}}}
		&\sum_{j=1}^{m}\sum_{i=1}^{n}\|\boldsymbol{\rm b}_j-\boldsymbol{\rm V}\boldsymbol{\rm W}^T\boldsymbol{\rm x}_i\|_2^2\boldsymbol{\rm A}_{ij} + \alpha\|\boldsymbol{\rm W}\|_{2,1}\\
		&s.t.\  \boldsymbol{\rm B}\in \{-1,1\}^{l\times m}, \boldsymbol{\rm V}^T\boldsymbol{\rm V} = \boldsymbol{\rm I}\\
	\end{aligned},
\end{equation}
where $\alpha$ is the balance parameter, $\boldsymbol{\rm W}\in {\rm R}^{d\times l}$ is the pairwise weight matrix; $\boldsymbol{\rm V}\in {\rm R}^{l\times l}$ is the rotation matrix. Although JSH effectively handles the pairwise projection from a high-dimensional space to the Hamming space, the semantics of instances are unknown and motivate our work reported in this paper.

\section{Jointly Personalized Sparse Hashing}
\label{Jointly Personalized Sparse Hashing}
In this section, we present the details of the proposed JPSH, including semantic-preserving Personalized Sparse Hashing (PSH), overall objective function of JPSH, discussions, optimization, the out-of-sample extension.

\subsection{Personalized Sparse Hashing}
The semantic structure can reflect the intrinsic geometric relationship across different data points within a specific category. The proposed PSH thinks that instances with similar semantic structures should be adaptively projected to the same Hamming space, and vice versa, as shown in Figure \ref{fig:1} (b). For this purpose, one simple yet effective way is to build different personalized subspaces \cite{weinberger2009feature} to capture different semantic structures. Retrieving label information is an expensive process in terms of time, labor and human expertise, and we define semantic structures in a high-dimensional space by pseudo labels produced by K-means. The instances with similar semantic information likely share the same anchor point. Therefore, we derive a set of anchor points $\boldsymbol{\rm C} = \{\boldsymbol{\rm c}_j\}_{j=1}^m$ as the training set of PSH and obtaining the following objective function:
\begin{equation}\label{eq4}
	\begin{aligned}
		\mathop{\min}_{\substack{\boldsymbol{\rm B},\boldsymbol{\rm P},\boldsymbol{\rm R}}}
		&\sum_{j = 1}^{m}\|\boldsymbol{\rm b}_j-\boldsymbol{\rm R}\boldsymbol{\rm P}_j^T\boldsymbol{\rm c}_j\|_2^2\\
		&s.t.\  \boldsymbol{\rm B}\in \{-1,1\}^{l\times m}, \boldsymbol{\rm R}^T\boldsymbol{\rm R} = \boldsymbol{\rm I}\\
	\end{aligned},
\end{equation}
where $\boldsymbol{\rm P}_j\in {\rm R}^{d\times l}$ is a personalized weight matrix for the anchor point $\boldsymbol{\rm c}_j$ and $\boldsymbol{\rm P}=\left[\boldsymbol{\rm P}_1;\boldsymbol{\rm P}_2;\dots;\boldsymbol{\rm P}_m\right]\in {\rm R}^{md\times l}$. $\boldsymbol{\rm R}\in {\rm R}^{l\times l}$ is a rotation matrix to minimize quantization errors, driven by ITQ \cite{gong2012iterative}.

Since the binarization operation for low-dimensional features $\boldsymbol{\rm R}\boldsymbol{\rm P}_j^T\boldsymbol{\rm c}_j$ and binary codes $\boldsymbol{\rm b}_j$ helps increase the information loss, we consider using the feature sparsity for personalized weight $\boldsymbol{\rm P}_j$ to select important features. Exclusive group lasso (EGL) \cite{kong2014exclusive,kong2016uncorrelated} encourages intra-cluster competition but discourages inter-cluster competition. Inspired by EGL we first impose a $l_{2,1}$-norm regularization on personalized weight $\boldsymbol{\rm P}_j$ for pursuing the sparsity of intra-cluster features. Then, we create a $l_2$-norm term to constrain the non-sparsity of inter-cluster features. This regularization term enables us to perform optimal feature selection for each cluster. However, moderating the personalized weight $\boldsymbol{\rm P}_j$ with a single anchor point $\boldsymbol{\rm c}_j$ may over-fit with a poor generalization ability. Therefore, we force the anchor point $\boldsymbol{\rm c}_j$ to lend certain strengths from its neighboring anchors to learn the personalized weight $\boldsymbol{\rm P}_j$, motivated by the network lasso penalty \cite{hallac2015network}. Finally, we re-write the loss function of PSH as follows:
\begin{equation}\label{eq5}
	\begin{aligned}
		\mathop{\min}_{\substack{\boldsymbol{\rm B},\boldsymbol{\rm P},\boldsymbol{\rm R}}}
		&\sum_{j = 1}^{m}\|\boldsymbol{\rm b}_j-\boldsymbol{\rm R}\boldsymbol{\rm P}_j^T\boldsymbol{\rm c}_j\|_2^2+\lambda_1\sum_{j=1}^{m}\|\boldsymbol{\rm P}_j\|_{2,1}^{2} + \lambda_2\sum_{i,j = 1}^{m}\boldsymbol{\rm S}_{ij}\|\boldsymbol{\rm P}_i-\boldsymbol{\rm P}_j\|_F\\
		&s.t.\  \boldsymbol{\rm B}\in \{-1,1\}^{l\times m}, \boldsymbol{\rm R}^T\boldsymbol{\rm R} = \boldsymbol{\rm I}\\
	\end{aligned},
\end{equation}
where $\lambda_1$ and $\lambda_2$ are balance parameters, $\boldsymbol{\rm S}\in {\rm R}^{m\times m}$ is the similarity matrix of anchors, and the $i$-th element of $\boldsymbol{\rm S}_{*j}$ is defined as follows:
\begin{equation}\label{eq6}
	\boldsymbol{\rm S}_{ij}=\left\{
	\begin{aligned}
		&{\rm exp}\left(-\frac{\|\boldsymbol{\rm c}_i-\boldsymbol{\rm c}_j\|_2^2}{\delta^2}\right)  ,{\rm if} \ \boldsymbol{\rm c}_i \in N_\psi(\boldsymbol{\rm c}_j) \ {\rm or} \ \boldsymbol{\rm c}_j \in N_\psi(\boldsymbol{\rm c}_i) \\
		&0 \quad \quad \quad \quad \quad  \quad  \quad \quad \ ,{\rm otherwise}
	\end{aligned},
	\right.
\end{equation}
where $\delta$ is a given parameter, and $N_\psi(\boldsymbol{\rm c}_i)$ is the index set of $\psi$ nearest neighbors of the anchor point $\boldsymbol{\rm c}_i$.

\subsection{Overall Objective Function of JPSH} 
As aforementioned, the PSH model shown in Eq.~\eqref{eq5} can effectively preserve particular attributes of each cluster from the high-dimensional space into the Hamming space. Instances within the same cluster can be projected by the same personalized weight matrix, resulting in directly maintaining the semantics of instances that correspond to intrinsic structures of data into the Hamming space. 

To further improve the representation capacity of binary codes, we also hope to preserve the local neighborhood structure within the cluster in the Hamming space. To this end, we maintain the pairwise similarity via a manifold-based hashing learning method. Since JSH \cite{lai2018jointly} can quantify the similarity between each instance and its neighboring anchor points, and greatly reduce the time complexity, we thereby joint Eqs.~\eqref{eq3} and~\eqref{eq5} to construct the Jointly Personalized Sparse Hashing (JPSH) model. The overall objective function of JPSH is defined as:
\begin{equation}\label{eq7}
	\begin{aligned}
		&\mathop{\min}_{\substack{\boldsymbol{\rm B},\boldsymbol{\rm P},\boldsymbol{\rm W},\\ \boldsymbol{\rm R},\boldsymbol{\rm V}}}
		\sum_{j = 1}^{m}\|\boldsymbol{\rm b}_j-\boldsymbol{\rm R}\boldsymbol{\rm P}_{j}^T\boldsymbol{\rm c}_j\|_2^2+\sum_{j=1}^{m}\sum_{i=1}^{n}\|\boldsymbol{\rm b}_j-\boldsymbol{\rm V}\boldsymbol{\rm W}^T\boldsymbol{\rm x}_i\|_2^2\boldsymbol{\rm A}_{ij}\\
		&\ \ \ \ +\lambda_1\sum_{j=1}^{m}\|\boldsymbol{\rm P}_{j}\|_{2,1}^{2}+\lambda_2\sum_{i,j = 1}^{m}\boldsymbol{\rm S}_{ij}\|\boldsymbol{\rm P}_{i}-\boldsymbol{\rm P}_{j}\|_F+\lambda_3\|\boldsymbol{\rm W}\|_{2,1} \\
		&\ \ \ \ \ \ \ \ \ s.t.\  \boldsymbol{\rm B}\in \{-1,1\}^{l\times m}, \boldsymbol{\rm R}^T\boldsymbol{\rm R} = \boldsymbol{\rm V}^T\boldsymbol{\rm V} = \boldsymbol{\rm I}\\
	\end{aligned}
\end{equation}
where $\lambda_3$ is the balance parameter for projected matrix $\boldsymbol{\rm W}$. 

\subsection{Discussions}
In the following, we will discuss how our proposed JPSH learns the binary representation and preserves discriminative information compared with other hashing methods from two aspects. 

On the one hand, compared to existing manifold-based hashing methods, such as SH and JSH, our proposed JPSH not only considers local pairwise similarity in the original feature space $\boldsymbol{\rm X}$, but also the intrinsic semantic similarity of the data. In particular, Eq.~\eqref{eq3} shows that JSH can effectively reduce the computational complexity using truncated similarity matrix $\boldsymbol{\rm A}$, and avoid information loss via $\|\boldsymbol{\rm W}\|_{2,1}$. Therefore, we maintain local neighborhood structures within each cluster by JSH, as the local structure information in the two clusters (purple points and blue points shown in the Figure \ref{fig:2} (b)). However, JSH has difficulty distinguishing instances with similar features but different semantic information. Comparing Eq.~\eqref{eq7} against Eq.~\eqref{eq3}, we learn a set of personalized weights $\boldsymbol{\rm P}_j \left(j = 1,2,\dots,m\right)$ for different anchors $\boldsymbol{\rm c}_j \left(j = 1,2,\dots,m\right)$, and employ the $l_{2,1}$-norm sparse constraint on each $\boldsymbol{\rm P}_j$. So, instances with the same pseudo-label can be adaptively mapped to the personalized subspace through the unique projection learning of personalized weights. Therefore, the proposed JPSH can avoid misrepresenting instances of similar features with different semantics by directly retaining the semantic relationship of the original feature space.

On the other hand, compared against those multi-stage hashing methods, such as PCA-ITQ and RSSH, we construct the seamless hash objective function Eq.~\eqref{eq7}. It can avoid suboptimal solutions generated by using multiple independent learning strategies. For example, PCA-ITQ first applies PCA to perform linear dimensionality reduction, and then uses an alternating algorithm for refining the initial orthogonal transformation to reduce quantization errors. RSSH preserves semantic relationships on three subproblems: multi-subspace learning, similarity matrix construction, and semantics preserved hashing. Different from the above methods, we construct personalized weight matrix $\boldsymbol{\rm P}$, pairwise weight matrix $\boldsymbol{\rm W}$ and two rotation matrices $\boldsymbol{\rm R}$ and $\boldsymbol{\rm V}$ in the seamless formulation Eq.~\eqref{eq7}. Thus, our proposed JPSH can significantly avoid the suboptimal solution, and obtain discriminative hash codes in binary representation learning.

\subsection{Optimization}
\label{optimization}
The optimization of Eq.~\eqref{eq7} is intractable and non-convex w.r.t all parameters simultaneously. Therefore, we adopt an alternating optimization algorithm to iteratively update all parameters till the algorithm converges to an acceptable settlement.

\textbf{Update Personalized Weight \texorpdfstring{$\boldsymbol{\rm P}$}{E2}}: First, we update the personalized weight $\boldsymbol{\rm P}$ when $\boldsymbol{\rm W}$, $\boldsymbol{\rm R}$, $\boldsymbol{\rm V}$ and $\boldsymbol{\rm B}$ are fixed. The subproblem of Eq.~\eqref{eq7} w.r.t $\boldsymbol{\rm P}$ is:
\begin{equation}\label{eq8}
	\begin{aligned}
		\mathop{\min}_{\substack{\boldsymbol{\rm P}}}
		&\sum_{j = 1}^{m}\|\boldsymbol{\rm b}_j-\boldsymbol{\rm R}\boldsymbol{\rm P}_j^T\boldsymbol{\rm c}_j\|_2^2 + \lambda_1\sum_{j=1}^{m}\|\boldsymbol{\rm P}_j\|_{2,1}^{2} + \lambda_2\sum_{i,j = 1}^{m}\boldsymbol{\rm S}_{ij}\|\boldsymbol{\rm P}_i-\boldsymbol{\rm P}_j\|_F \\
		&s.t.\  \boldsymbol{\rm B}\in \{-1,1\}^{l\times m},\boldsymbol{\rm R}^T\boldsymbol{\rm R} = \boldsymbol{\rm I}\\
	\end{aligned}.
\end{equation}

For the convenience, we devide Eq.~\eqref{eq8} into three functions: $F_1(\boldsymbol{\rm P})=\sum_{j = 1}^{m}\|\boldsymbol{\rm b}_j-\boldsymbol{\rm R}\boldsymbol{\rm P}_j^T\boldsymbol{\rm c}_j\|_2^2$, $F_2(\boldsymbol{\rm P})=\lambda_1\sum_{j=1}^{m}\|\boldsymbol{\rm P}_j\|_{2,1}^{2}$, $F_3(\boldsymbol{\rm P})=\lambda_2\sum_{i,j = 1}^{m}\boldsymbol{\rm S}_{ij}\|\boldsymbol{\rm P}_i-\boldsymbol{\rm P}_j\|_F$. We denote $\boldsymbol{\rm Y}={\rm diag}\left(\boldsymbol{\rm c}_1,\boldsymbol{\rm c}_2,\dots,\boldsymbol{\rm c}_m\right)$, and $\boldsymbol{\rm Y}\in {\rm R}^{md\times m}$, where ${\rm diag}\left(\cdot \right)$ is a diagonal function. Since $\boldsymbol{\rm R}\boldsymbol{\rm R}^T=\boldsymbol{\rm I}$, the derivative of $F_1(\boldsymbol{\rm P})$ w.r.t  $\boldsymbol{\rm P}$ can be computed as follows:
\begin{equation}\label{eq9}
	\begin{aligned}
		\frac{\partial F_1(\boldsymbol{\rm P})}{\partial \boldsymbol{\rm P}} = -2\boldsymbol{\rm Y}\boldsymbol{\rm B}^T\boldsymbol{\rm R}+2\boldsymbol{\rm Y}\boldsymbol{\rm Y}^T\boldsymbol{\rm P}\\
	\end{aligned}.
\end{equation}

According to Refs. \cite{li2017toward, li2018unsupervised}, we have the derivative of $F_2(\boldsymbol{\rm P})$ w.r.t $\boldsymbol{\rm P}$ as follows:
\begin{equation}\label{eq10}
	\begin{aligned}
		\frac{\partial F_2(\boldsymbol{\rm P})}{\partial \boldsymbol{\rm P}} = 2\lambda_1\boldsymbol{\rm K}\boldsymbol{\rm P}\\
	\end{aligned},
\end{equation}
where $\boldsymbol{\rm K}\in {\rm R}^{md\times md}$ is a diagonal matrix, and the $i$-th diagonal element can be defined as:
\begin{equation}\label{eq11}
	\begin{aligned}
		\boldsymbol{\rm K}_{ii} = \sum_{j = 1}^{m}\frac{\boldsymbol{\mathbb{I}}_{ij}\|\boldsymbol{\rm P}_j\|_{2,1}}{\|\boldsymbol{\rm P}^{i*}\|_2+\epsilon}\\
	\end{aligned},
\end{equation}
where $\epsilon$ is a very small constant to ensure the derivative solvable, $\boldsymbol{\mathbb{I}}\left(\cdot \right)$ is an indicator function, and $\boldsymbol{\mathbb{I}}_{ij}=1$ if $\boldsymbol{\rm P}^{i*}$ belongs to $\boldsymbol{\rm P}_j$, otherwise $\boldsymbol{\mathbb{I}}_{ij}=0$.

The derivative of $F_3(\boldsymbol{\rm P})$ w.r.t. $\boldsymbol{\rm P}$ is computed as follows:
\begin{equation}\label{eq12}
	\begin{aligned}
		\frac{\partial F_3(\boldsymbol{\rm P})}{\partial \boldsymbol{\rm P}} = 2\lambda_2\left(\boldsymbol{\rm G}\otimes \boldsymbol{\rm I}_d\right)\boldsymbol{\rm P}\\
	\end{aligned}.
\end{equation}
where $\boldsymbol{\rm I}_d\in {\rm R}^{d\times d}$ is an identity matrix. $\boldsymbol{\rm G}\in {\rm R}^{m\times m}$ is square matrix, and the $i$-th element of $\boldsymbol{\rm G}_{*j}$ can be expressed as:
\begin{equation}\label{eq13}
	\boldsymbol{\rm G}_{ij}=\left\{
	\begin{aligned}
		&\sum_{k=1}^{m}\frac{\boldsymbol{\rm S}_{ik}}{||\boldsymbol{\rm P}_i-\boldsymbol{\rm P}_k||_F+\epsilon}-\frac{\boldsymbol{\rm S}_{ij}}{||\boldsymbol{\rm P}_i-\boldsymbol{\rm P}_j||_F+\epsilon} & \left(i=j\right)\\
		&\frac{-\boldsymbol{\rm S}_{ij}}{||\boldsymbol{\rm P}_i-\boldsymbol{\rm P}_j||_F+\epsilon} & \left(i\neq j\right)
	\end{aligned}.
	\right.
\end{equation}

Combining $\partial F_1(\boldsymbol{\rm P})/\partial \boldsymbol{\rm P}$, $\partial F_2(\boldsymbol{\rm P})/\partial \boldsymbol{\rm P}$ and $\partial F_3(\boldsymbol{\rm P})/\partial \boldsymbol{\rm P}$ together, and setting it to be zero, we have a closed-form solution of the personalized weight $\boldsymbol{\rm P}$ as follows:
\begin{equation}\label{eq14}
	\begin{aligned}
		\boldsymbol{\rm P} = \left(\lambda_1\boldsymbol{\rm K}+\lambda_2\left(\boldsymbol{\rm G}\otimes \boldsymbol{\rm I}_d\right)+\boldsymbol{\rm Y}\boldsymbol{\rm Y}^T\right)^{-1}\boldsymbol{\rm Y}\boldsymbol{\rm B}^T\boldsymbol{\rm R}
	\end{aligned},
\end{equation}

\textbf{Update Pairwise Weight \texorpdfstring{$\boldsymbol{\rm W}$}{E1}}: We attempt to update the pairwise weight $\boldsymbol{\rm W}$ when other four parameters are fixed. Removing terms that are irrelevant to $\boldsymbol{\rm W}$, we have the following function as:
\begin{equation}\label{eq15}
	\begin{aligned}
		\mathop{\min}_{\substack{\boldsymbol{\rm W}}}
		&\sum_{j=1}^{m}\sum_{i=1}^{n}\|\boldsymbol{\rm b}_j-\boldsymbol{\rm V}\boldsymbol{\rm W}^T\boldsymbol{\rm x}_i\|_2^2\boldsymbol{\rm A}_{ij} +\lambda_3\|\boldsymbol{\rm W}\|_{2,1}\\
		&s.t.\  \boldsymbol{\rm B}\in \{-1,1\}^{l\times m}, \boldsymbol{\rm V}^T\boldsymbol{\rm V} = \boldsymbol{\rm I}\\
	\end{aligned}.
\end{equation}
According to Ref. \cite{lai2018jointly}, since the $\sum_i \boldsymbol{\rm A}_{ij} = 1$ and $\boldsymbol{\rm V}\boldsymbol{\rm V}^T=\boldsymbol{\rm I}$, the term between $\boldsymbol{\rm X}$ and $\boldsymbol{\rm X}^T$ is an identity matrix. So, the closed-form solution of $\boldsymbol{\rm W}$ is:
\begin{equation}\label{eq16}
	\begin{aligned}
		\boldsymbol{\rm W} = \left(\lambda_3\boldsymbol{\rm Q}+\boldsymbol{\rm X}\boldsymbol{\rm X}^T\right)^{-1}\boldsymbol{\rm X}\boldsymbol{\rm A}\boldsymbol{\rm B}^T\boldsymbol{\rm V}
	\end{aligned},
\end{equation}
where $\boldsymbol{\rm Q}\in {\rm R}^{d\times d}$ is a diagonal matrix, and the $i$-th element of $\boldsymbol{\rm Q}_{i*}$ is:
\begin{equation}\label{eq17}
	\begin{aligned}
		\boldsymbol{\rm Q}_{ii} = \frac{1}{2\|\boldsymbol{\rm w}_{i*}\|_2}
	\end{aligned},
\end{equation}
where $\boldsymbol{\rm w}_{i*}$ is the $i$-th row of the pairwise weight $\boldsymbol{\rm W}$.

\textbf{Update Rotation Matrix \texorpdfstring{$\boldsymbol{\rm R}$}{E4}}: To compute $\boldsymbol{\rm R}$, we fix $\boldsymbol{\rm P}$, $\boldsymbol{\rm W}$, $\boldsymbol{\rm V}$ and $\boldsymbol{\rm B}$, and then solve the following maximization problem:
\begin{equation}\label{eq18}
	\begin{aligned}
		\mathop{\max}_{\substack{\boldsymbol{\rm R}}}\ {\rm tr}\left(\boldsymbol{\rm R}\boldsymbol{\rm P}^T\boldsymbol{\rm Y}\boldsymbol{\rm B}^T\right)\\
	\end{aligned},
\end{equation}
According to \cite{zou2006sparse, li2017large}, we compute $\boldsymbol{\rm P}^T\boldsymbol{\rm Y}\boldsymbol{\rm B}^T=\boldsymbol{\rm \hat{U}}_1\boldsymbol{\rm D}_1\boldsymbol{\rm \hat{V}}_1^T$ by Singular Value Decomposition (SVD), and the updating scheme of $\boldsymbol{\rm R}$ is described below:
\begin{equation}\label{eq19}
	\begin{aligned}
		\boldsymbol{\rm R} = \boldsymbol{\rm \hat{V}}_1\boldsymbol{\rm \hat{U}}_1^T
	\end{aligned}.
\end{equation}

\textbf{Update Rotation Matrix \texorpdfstring{$\boldsymbol{\rm V}$}{E3}}: To derive the solution of $\boldsymbol{\rm V}$, we fix the other four parameters and re-form the maximization problem as follows:
\begin{equation}\label{eq20}
	\begin{aligned}
		\mathop{\max}_{\substack{\boldsymbol{\rm V}}}\ {\rm tr}\left(\boldsymbol{\rm V}\boldsymbol{\rm W}^T\boldsymbol{\rm X}\boldsymbol{\rm A}\boldsymbol{\rm B}^T\right)\\
	\end{aligned}.
\end{equation}
We use SVD to solve the problem $\boldsymbol{\rm W}^T\boldsymbol{\rm X}\boldsymbol{\rm A}\boldsymbol{\rm B}^T=\boldsymbol{\rm \hat{U}}_2\boldsymbol{\rm D}_2\boldsymbol{\rm \hat{V}}_2^T$, and then retain the solution of $\boldsymbol{\rm V}$ as follows:
\begin{equation}\label{eq21}
	\begin{aligned}
		\boldsymbol{\rm V} = \boldsymbol{\rm \hat{V}}_2\boldsymbol{\rm \hat{U}}_2^T
	\end{aligned}.
\end{equation}

\textbf{Update Binary Codes \texorpdfstring{$\boldsymbol{\rm B}$}{E5}}: Finally, we update the parameter $\boldsymbol{\rm B}$ when others have been fixed and solve the optimization problem as follows:
\begin{equation}\label{eq22}
	\begin{aligned}
		\mathop{\max}_{\substack{\boldsymbol{\rm B}}}\ {\rm tr}\left(\boldsymbol{\rm R}\boldsymbol{\rm P}^T\boldsymbol{\rm Y}\boldsymbol{\rm B}^T+\boldsymbol{\rm V}\boldsymbol{\rm W}^T\boldsymbol{\rm X}\boldsymbol{\rm A}\boldsymbol{\rm B}^T\right)\\
	\end{aligned}
\end{equation}
Thus, the solution to $\boldsymbol{\rm B}$ is:
\begin{equation}\label{eq23}
	\begin{aligned}
		\boldsymbol{\rm B} = \mathrm{sgn}\left(\boldsymbol{\rm R}\boldsymbol{\rm P}^T\boldsymbol{\rm Y}+\boldsymbol{\rm V}\boldsymbol{\rm W}^T\boldsymbol{\rm X}\boldsymbol{\rm A}\right)
	\end{aligned}
\end{equation}
where $\mathrm{sgn}\left(\cdot\right)$ is a sign function.

With the above handling, the pseudo codes of our proposed JPSH method is summarized in Algorithm \ref{alg:1}.
\begin{algorithm}
	\renewcommand{\algorithmicrequire}{\textbf{Input:}}
	\renewcommand{\algorithmicensure}{\textbf{Output:}}
	\caption{Jointly Personalized Sparse Hashing (JPSH)}\label{alg:1}
	\begin{algorithmic}[1]
		\REQUIRE Training set $\boldsymbol{\rm X}$, balance parameters $\lambda_1$, $\lambda_2$, $\lambda_3$, the number of anchors $m$, $k$ neighboring anchor points, the length of hash codes $l$, and the iteration time $T$.
		\ENSURE Binary codes $\boldsymbol{\rm B}$, personalized weight $\boldsymbol{\rm P}$, pairwise weight $\boldsymbol{\rm W}$, and two rotation matrices $\boldsymbol{\rm R}$ and $\boldsymbol{\rm V}$.
		\STATE Obtain $m$ anchor points by the K-means method.
		\STATE Compute truncated similarity matrix $\boldsymbol{\rm A}$ and similarity matrix $\boldsymbol{\rm S}$ by Eqs.~\eqref{eq2} and ~\eqref{eq6}, respectively.
		\STATE Initialize $\boldsymbol{\rm R}$ and $\boldsymbol{\rm V}$ as orthogonal matrices, $\boldsymbol{\rm D}$ and $\boldsymbol{\rm Q}$ as identity matrices, and $\boldsymbol{\rm B}$ as a random binary matrix.
		\WHILE {Loop until converge or reach iteration time $T$}
		\STATE Update $\boldsymbol{\rm P}$ by Eq.~\eqref{eq14};
		\STATE Update $\boldsymbol{\rm W}$ by Eq.~\eqref{eq16};
		\STATE Update $\boldsymbol{\rm Q}$ by Eq.~\eqref{eq17};
		\STATE Update $\boldsymbol{\rm R}$ by Eq.~\eqref{eq19};
		\STATE Update $\boldsymbol{\rm V}$ by Eq.~\eqref{eq21};
		\STATE Update $\boldsymbol{\rm B}$ by Eq.~\eqref{eq23};
		\ENDWHILE
		\STATE \textbf{return} $\boldsymbol{\rm B}$, $\boldsymbol{\rm P}$, $\boldsymbol{\rm W}$, $\boldsymbol{\rm R}$ and $\boldsymbol{\rm V}$.
	\end{algorithmic}
\end{algorithm}

\subsection{Out-of-Sample Extension}
After having optimized the overall objective function of JPSH, we perform similarity prediction based on four parameters $\boldsymbol{\rm P}$, $\boldsymbol{\rm W}$, $\boldsymbol{\rm R}$ and $\boldsymbol{\rm V}$. More specifically, for each query instance $\boldsymbol{\rm \hat{x}}$, we first find the personalized weight $\boldsymbol{\rm P}_j$ corresponding to the anchor point $\boldsymbol{\rm c}_j$ using the minimum Euclidean distance, for the query instance $\boldsymbol{\rm \hat{x}}$. Then, we joint the personalized weight $\boldsymbol{\rm P}_j$, pairwise weight $\boldsymbol{\rm W}$, and two rotation matrices $\boldsymbol{\rm R}$ and $\boldsymbol{\rm V}$ to learn discriminative binary codes $\boldsymbol{\rm \hat{b}}$, that is: $\boldsymbol{\rm \hat{b}}=\mathrm{sgn}\left(\boldsymbol{\rm R}\boldsymbol{\rm P}_j^T\boldsymbol{\rm c}_j+\boldsymbol{\rm V}\boldsymbol{\rm W}^T\boldsymbol{\rm \hat{x}}\right)$.

Observing $\boldsymbol{\rm \hat{b}}=\mathrm{sgn}\left(\boldsymbol{\rm R}\boldsymbol{\rm P}_j^T\boldsymbol{\rm c}_j+\boldsymbol{\rm V}\boldsymbol{\rm W}^T\boldsymbol{\rm \hat{x}}\right)$, we can find that each hash code $\boldsymbol{\rm \hat{b}}$ is constrained by $\boldsymbol{\rm R}\boldsymbol{\rm P}_j^T\boldsymbol{\rm c}_j$ and $\boldsymbol{\rm V}\boldsymbol{\rm W}^T\boldsymbol{\rm \hat{x}}$. The first term $\boldsymbol{\rm R}\boldsymbol{\rm P}_j^T\boldsymbol{\rm c}_j$ transforms the anchor $\boldsymbol{\rm c}_j$ corresponding to $\boldsymbol{\rm \hat{x}}$ by the personalized weight matrix $\boldsymbol{\rm P}_{j}$. Since the personalized weight matrices are different for different clusters, the hash codes $\boldsymbol{\rm \hat{x}}$ generated by the first trem have the semantic separability. From Eq.~\eqref{eq7}, we can observe that the pairwise weight matrix $\boldsymbol{\rm W}$ is produced under the constraints of pairwise similarity matrix $\boldsymbol{\rm A}$. Thus, the second trem $\boldsymbol{\rm V}\boldsymbol{\rm W}^T\boldsymbol{\rm \hat{x}}$ reflects the local neighborhood structures among data. Integrating the first term $\boldsymbol{\rm R}\boldsymbol{\rm P}_j^T\boldsymbol{\rm c}_j$ and the second term $\boldsymbol{\rm V}\boldsymbol{\rm W}^T\boldsymbol{\rm \hat{x}}$ into the program of producing $\boldsymbol{\rm \hat{b}}$, we can obtain the binary codes fused the semantic and pairwise similarities.

\section{Experimental WorkS}
\label{Experimental Work}
To verify the effectiveness of the proposed binary representation algorithm JPSH, we use it in the task of similarity search. We conduct extensive experiments on four widely-used benchmark image datasets, and compare with several state-of-the-art hash algorithms.

\subsection{Datasets}
\begin{figure}[h]
	\centering
	\includegraphics[width=\linewidth]{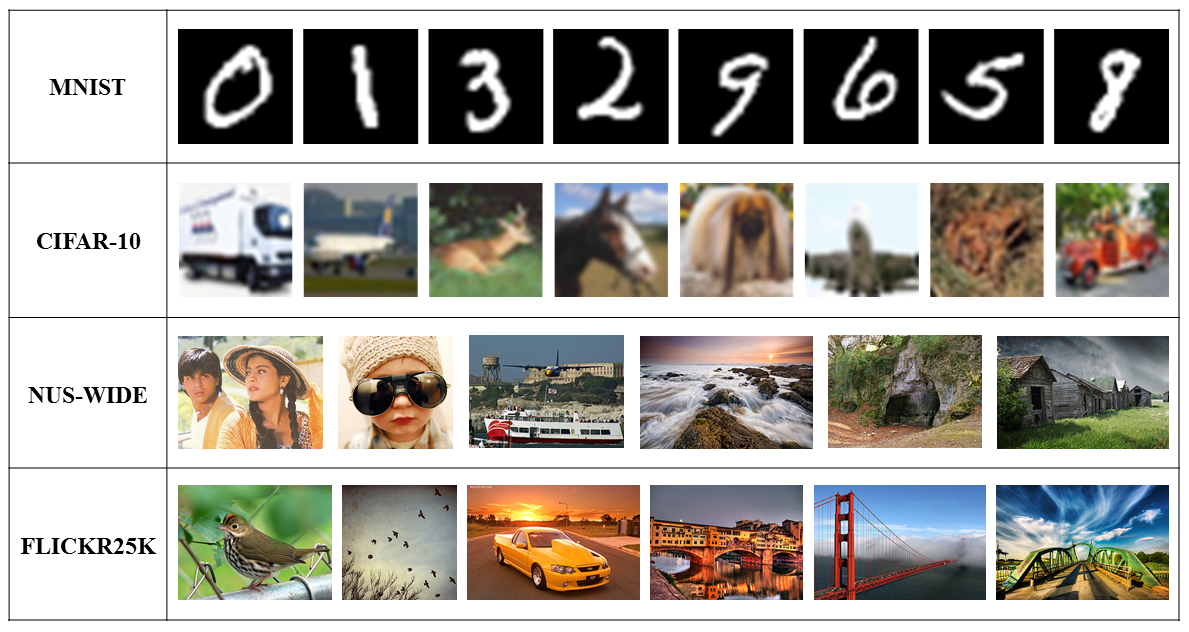}
	\caption{Example images of the used datasets, including MNIST, CIFAR-10, NUS-WIDE and FLICKR25K.}
	\label{fig:3}
\end{figure}

The proposed method and the comparative ones are evaluated on the following four datasets, and some example images of each dataset are shown in Figure~\ref{fig:3}. \textbf{MNIST} \cite{lecun1998gradient} contains 70,000 handwritten digit images from 10 classes in total. Each image is re-shaped to a 784-dimensional feature vector. We construct the testing set by randomly selecting 100 images per class, and the remaining images are composed as the training set \cite{lin2018supervised}. \textbf{CIFAR-10} \cite{krizhevsky2009learning} consists of 60,000 tiny images of 10 classes. Following \cite{hu2018discrete}, 512-dimensional GIST features \cite{oliva2001modeling} are retrieved. We randomly select 100 images per class to form the testing set, and the remaining 5900 images per class form the training set. \textbf{NUS-WIDE} \cite{chua2009nus} is a real-world dataset containing 269,648 images related to 81 ground truth concepts. We pick the most 21 frequent concepts for evaluation. For each category, 100 images are randomly sampled to form the testing set, and 100,000 images from the remaining images form the training set. 4096-dimensional CNN features \cite{krizhevsky2017imagenet} are firstly extracted, and then 1000-dimensional PCA features are generated to represent each image. \textbf{FLICKR25K} \cite{huiskes2008mir} is also a real-world dataset including 25,000 images of 24 categories. We randomly select 1000 images to form the testing set, and 19,015 images for the training set \cite{9248040}. Following the setting of NUS-WIDE, we extract 1000-dimensional PCA features to represent each image.  

\subsection{Settings}
\label{Settings}
For the proposed JPSH, the test range of parameters $\lambda_1$, $\lambda_2$ and $\lambda_3$ is $\{0, 1, \dots, 10^5, 10^6\}$, the test range of parameter $k$ is $\left[3, 5, 7,10, 20\right]$, and the test range of parameter $m$ is $\left[100, 200, \dots, 1000\right]$. Based on the test results, we empirically set $k=7$, $m=800$, $\lambda_1=\lambda_2=1$ and $\lambda_3=10$ on MNIST and CIFAR-10, and set $k=5$, $\lambda_1=\lambda_2=10$ and $\lambda_3=10^5$ on NUS-WIDE and FLICKR25K, while $m=1000$ for NUS-WIDE, and $m=500$ for FLICKR25K. In addition, we set $\psi = 7$ in Eq.~\eqref{eq6} for MNIST and CIFAR-10 datasets, and set $\psi = 5$ for FLICKR25K and NUS-WIDE datasets. We test the system with different lengths of codes ranged in the set of $\left[8, 16, 32, 64, 96, 128\right]$. As for the evaluation, five standard evaluation metrics, i.e. mean Average Precision (mAP), the mean average precision of the top 100 testing samples (Pre@100), Precision-Recall curve, precision vs. top-N positions (Precision@N) curve, and recall vs. top-N positions (Recall@N) curve, are used to evaluate the performance. All results shown here are produced with Hamming radius 2, being the average of 10 run times. The baseline methods used in this evaluation are under no domain adaption assumption. The baseline methods consist of 10 state-of-the-art unsupervised hashing methods: LSH \cite{datar2004locality}, SH \cite{weiss2008spectral}, PCA-ITQ \cite{gong2012iterative}, SP \cite{xia2015sparse}, SGH \cite{jiang2015scalable}, OEH \cite{liu2016towards}, OCH \cite{liu2018ordinal}, JSH \cite{lai2018jointly}, CH \cite{weng2020concatenation} and RSSH \cite{tian2020unsupervised}. All hashing methods are examined using Matlab on a PC with 3.6GHz and 64G RAM.

\subsection{Ablation Studies}
To analyze the roles of pairwise and semantic similarities played in JPSH, we perform ablation studies on the CIFAR-10 and FLICKR25K datasets. The experimental results are depicted in Table~\ref{tab:1}. From this table, we have three clear observations: Firstly, JPSH achieves better performance than JSH and PSH. The reason is that JPSH learns and creates a seamless hash function, jointly considering pairwise and semantic similarities. Compared to JSH, the proposed JPSH considers further the semantics of instances corresponding to the intrinsic geometric structures. In addition, the JPSH retains the superior performance of the traditional manifold hashing method JSH comparsed with the PSH. Secondly, closely looking at JSH and PSH, we observe that PSH performs worse than JSH. One possible reason is that pseudo labels determined by K-means cannot accurately represent the semantic information of the high-dimensional data, comparing to the ground truths. Third, only taking PSH into consideration, we discover that PSH can achieve decent performance, especially on real-world FLICKR25K dataset. This verifies that the proposed PSH can retain the semantic similarity of the high-dimensional space into the Hamming space in a certain extent, without any manually tagged labels. Overall, these well-designed studies exhibit the effectiveness of combining JSH-based pairwise and PSH-based semantic similarity to a seamless formulation. 

\begin{table}[h]
	\centering
	\caption{mAP and Pre@100 of JSH, PSH and JPSH on the CIFAR-10 and FLICKR25K datasets with 16 bits. Note that the best results are in bold and the second-best results are underlined.}
	\label{tab:1}
	\begin{tabular}{|c||c|c|c|c|c|c|}
		\hline
		\multirow{2}{*}{Datasets} &\multicolumn{2}{c|}{CIFAR-10} &\multicolumn{2}{c|}{FLICKR25K} \\
		\cline{2-5}
		&mAP &Pre@100 &mAP &Pre@100 \\
		\hline
		JSH\cite{lai2018jointly}   &\underline{0.1561} &\underline{0.3001}  &\underline{0.6667} &\underline{0.8909}  \\
		PSH   &0.1133 &0.1613  &0.5632 &0.8316  \\
		JPSH  &\textbf{0.1606} &\textbf{0.3275}  &\textbf{0.6936} &\textbf{0.9176}  \\
		\hline	
	\end{tabular}
\end{table}
\begin{table}[h]
	\centering
	\caption{mAP and Pre@100 for the JPSH and baselines on the MNIST, CIFAR-10, NUS-WIDE and FLICKR25K datasets. Note that the best results are in bold and the second-best results are underlined.}
	\label{tab:2}
	\resizebox{\textwidth}{!}{
		\begin{tabular}{|c||c|c|c|c|c|c|c|c|c|c|c|c|c|}
			\hline
			Datasets &Metrics &Bits &LSH\cite{datar2004locality} &SH\cite{weiss2008spectral} &PCA-ITQ\cite{gong2012iterative} &SP\cite{xia2015sparse} &SGH\cite{jiang2015scalable} &OEH\cite{liu2016towards} &OCH\cite{liu2018ordinal} &JSH\cite{lai2018jointly} &CH\cite{weng2020concatenation} &RSSH\cite{tian2020unsupervised} &JPSH\\
			\hline \hline
			\multirow{12}{*}{\rotatebox{90}{MNIST}}
			&\multirow{6}{*}{\rotatebox{90}{mAP}}		
			&8    &0.1808 &0.2651 &0.3589 &0.3404 &0.3111 &0.2920 &0.2376 &0.3204 &0.3030 &\underline{0.3940} &\textbf{0.4332}\\
			&&16  &0.2094 &0.2661 &0.4042 &0.3780 &0.3526 &0.3100 &0.3073 &0.4066 &0.3688 &\underline{0.4128} &\textbf{0.5561}\\
			&&32  &0.2633 &0.2606 &0.4378 &0.4178 &0.3898 &0.3497 &0.3727 &\underline{0.5118} &0.4317 &0.4183 &\textbf{0.6184}\\
			&&64  &0.3199 &0.2400 &0.4490 &0.4401 &0.4180 &0.3727 &0.4107 &\underline{0.5613} &0.4112 &0.4204 &\textbf{0.6548}\\
			&&96  &0.3592 &0.2367 &0.4596 &0.4472 &0.4282 &0.3889 &0.4253 &\underline{0.5675} &0.4318 &0.4222 &\textbf{0.6666}\\
			&&128 &0.3757 &0.2330 &0.4627 &0.4510 &0.4358 &0.4014 &0.4339 &\underline{0.5758} &0.4401 &0.4245 &\textbf{0.6659}\\
			\cline{2-14}
			&\multirow{6}{*}{\rotatebox{90}{Pre@100}}
			&8    &0.4372 &0.4859 &0.7165 &0.7057 &0.6835 &0.6040 &0.5581 &0.7213 &0.6644 &\underline{0.7843} &\textbf{0.8635}\\
			&&16  &0.4871 &0.4904 &0.7999 &0.7630 &0.7300 &0.6523 &0.6786 &\underline{0.8112} &0.7424 &0.7122 &\textbf{0.9396}\\
			&&32  &0.6102 &0.4732 &0.8468 &0.8169 &0.7828 &0.7233 &0.7461 &\underline{0.8954} &0.7954 &0.6795 &\textbf{0.9614}\\
			&&64  &0.7010 &0.4416 &0.8533 &0.8451 &0.8004 &0.7470 &0.7794 &\underline{0.9279} &0.7641 &0.6750 &\textbf{0.9619}\\
			&&96  &0.7565 &0.4379 &0.8685 &0.8524 &0.8111 &0.7915 &0.7931 &\underline{0.9370} &0.7914 &0.6737 &\textbf{0.9646}\\
			&&128 &0.7832 &0.4370 &0.8656 &0.8501 &0.8139 &0.7945 &0.8026 &\underline{0.9447} &0.8072 &0.6768 &\textbf{0.9656}\\
			\hline \hline
			\multirow{12}{*}{\rotatebox{90}{CIFAR-10}}
			&\multirow{6}{*}{\rotatebox{90}{mAP}}
			&8    &0.1102 &0.1232 &\textbf{0.1421} &0.1234 &0.1382 &0.1408 &0.1314 &0.1290 &\underline{0.1413} &0.1342 &0.1380\\
			&&16  &0.1134 &0.1266 &0.1472 &0.1265 &0.1497 &0.1489 &0.1494 &\underline{0.1561} &0.1447 &0.1428 &\textbf{0.1606}\\
			&&32  &0.1230 &0.1252 &0.1561 &0.1281 &0.1592 &0.1561 &\underline{0.1618} &0.1607 &0.1560 &0.1485 &\textbf{0.1812}\\
			&&64  &0.1306 &0.1254 &0.1485 &0.1268 &0.1684 &0.1646 &0.1705 &\underline{0.1785} &0.1658 &0.1533 &\textbf{0.1848}\\
			&&96  &0.1334 &0.1250 &0.1547 &0.1308 &0.1713 &0.1671 &0.1750 &\underline{0.1845} &0.1713 &0.1544 &\textbf{0.1925}\\
			&&128 &0.1373 &0.1249 &0.1576 &0.1333 &0.1734 &0.1694 &0.1765 &\underline{0.1865} &0.1738 &0.1574 &\textbf{0.1895}\\
			\cline{2-14}
			&\multirow{6}{*}{\rotatebox{90}{Pre@100}}
			&8    &0.1517 &0.1899 &\textbf{0.2612} &0.2019 &0.2382 &0.2486 &0.2157 &0.2187 &\underline{0.2601} &0.2138 &0.2413\\
			&&16  &0.1583 &0.1906 &0.2730 &0.1961 &0.2539 &0.2652 &0.2661 &\underline{0.3001} &0.2740 &0.2443 &\textbf{0.3275}\\
			&&32  &0.1915 &0.1760 &0.3063 &0.1979 &0.2839 &0.2876 &0.2989 &\underline{0.3313} &0.2894 &0.2626 &\textbf{0.3672}\\
			&&64  &0.2103 &0.1771 &0.2720 &0.1942 &0.3171 &0.3118 &0.3244 &\underline{0.3657} &0.3154 &0.2819 &\textbf{0.3910}\\
			&&96  &0.2120 &0.1748 &0.2988 &0.2047 &0.3263 &0.3177 &0.3371 &\underline{0.3765} &0.3318 &0.2828 &\textbf{0.4109}\\
			&&128 &0.2235 &0.1754 &0.3083 &0.2047 &0.3319 &0.3247 &0.3424 &\underline{0.3846} &0.3351 &0.2949 &\textbf{0.4015}\\
			\hline \hline
			\multirow{12}{*}{\rotatebox{90}{NUS-WIDE}}
			&\multirow{6}{*}{\rotatebox{90}{mAP}}
			&8    &0.3427 &0.4293 &\underline{0.5260} &0.5063 &0.4725 &0.4864 &0.4152 &0.4425 &0.4321 &0.3475 &\textbf{0.5521}\\
			&&16  &0.3598 &0.4067 &0.5243 &0.5032 &0.4600 &0.4780 &0.4672 &\underline{0.5511} &0.4712 &0.3614 &\textbf{0.6027}\\
			&&32  &0.3873 &0.3830 &0.5230 &0.5055 &0.4601 &0.4844 &0.5046 &\underline{0.5876} &0.4783 &0.3952 &\textbf{0.6200}\\
			&&64  &0.4138 &0.3689 &0.5286 &0.5133 &0.4815 &0.4903 &0.5350 &\underline{0.6138} &0.4758 &0.4126 &\textbf{0.6271}\\
			&&96  &0.4346 &0.3674 &0.5344 &0.5189 &0.4971 &0.4964 &0.5431 &\underline{0.6169} &0.4855 &0.4164 &\textbf{0.6293}\\
			&&128 &0.4479 &0.3769 &0.5397 &0.5224 &0.5059 &0.5019 &0.5487 &\underline{0.6224} &0.4947 &0.4211 &\textbf{0.6340}\\
			\cline{2-14}
			&\multirow{6}{*}{\rotatebox{90}{Pre@100}}
			&8    &0.6660 &0.7121 &\underline{0.8258} &0.8173 &0.7485 &0.8033 &0.7479 &0.7588 &0.7529 &0.6644 &\textbf{0.8313}\\
			&&16  &0.6875 &0.6998 &0.8243 &0.8142 &0.7408 &0.7991 &0.7805 &\underline{0.8420} &0.7796 &0.6752 &\textbf{0.8636}\\
			&&32  &0.7144 &0.6805 &0.8240 &0.8105 &0.7525 &0.8026 &0.8107 &\underline{0.8695} &0.7795 &0.6935 &\textbf{0.8762}\\
			&&64  &0.7357 &0.6809 &0.8265 &0.8179 &0.7729 &0.8018 &0.8342 &\textbf{0.8929} &0.8024 &0.6983 &\underline{0.8841}\\
			&&96  &0.7578 &0.6865 &0.8322 &0.8225 &0.7862 &0.8001 &0.8427 &\textbf{0.8877} &0.8026 &0.6947 &\underline{0.8860}\\
			&&128 &0.7638 &0.6986 &0.8362 &0.8255 &0.7935 &0.8117 &0.8439 &\textbf{0.8948} &0.8060 &0.6979 &\underline{0.8885}\\
			\hline \hline	
			\multirow{12}{*}{\rotatebox{90}{FLICKR25K}}
			&\multirow{6}{*}{\rotatebox{90}{mAP}}
			&8    &0.5802 &0.6321 &\underline{0.6949} &0.6828 &0.6539 &0.6653 &0.6360 &0.6206 &0.6711 &0.5674 &\textbf{0.6790}\\
			&&16  &0.5927 &0.6174 &\underline{0.6911} &0.6818 &0.6476 &0.6677 &0.6606 &0.6667 &0.6736 &0.5792 &\textbf{0.6936}\\
			&&32  &0.6100 &0.6023 &\underline{0.6907} &0.6819 &0.6539 &0.6698 &0.6859 &0.6875 &0.6735 &0.6028 &\textbf{0.7138}\\
			&&64  &0.6246 &0.5923 &0.6909 &0.6834 &0.6653 &0.6687 &\underline{0.7017} &0.7011 &0.6819 &0.6420 &\textbf{0.7253}\\
			&&96  &0.6403 &0.5928 &0.6923 &0.6864 &0.6742 &0.6759 &\underline{0.7070} &0.7042 &0.6838 &0.6521 &\textbf{0.7313}\\
			&&128 &0.6443 &0.5928 &0.6950 &0.6884 &0.6805 &0.6799 &\underline{0.7115} &0.7085 &0.6866 &0.6546 &\textbf{0.7331}\\
			\cline{2-14}
			&\multirow{6}{*}{\rotatebox{90}{Pre@100}}
			&8    &0.8417 &0.8701 &\underline{0.9079} &0.9027 &0.8874 &0.8991 &0.8773 &0.8603 &0.8959 &0.8361 &\textbf{0.9119}\\
			&&16  &0.8487 &0.8633 &\underline{0.9055} &0.9004 &0.8754 &0.8926 &0.8920 &0.8909 &0.8930 &0.8398 &\textbf{0.9176}\\
			&&32  &0.8587 &0.8590 &0.9039 &0.8998 &0.8803 &0.8941 &0.9057 &\underline{0.9077} &0.8938 &0.8482 &\textbf{0.9269}\\
			&&64  &0.8684 &0.8561 &0.9036 &0.8992 &0.8836 &0.8930 &0.9127 &\underline{0.9142} &0.9008 &0.8652 &\textbf{0.9291}\\
			&&96  &0.8745 &0.8565 &0.9049 &0.9014 &0.8892 &0.8957 &0.9151 &\underline{0.9167} &0.9029 &0.8688 &\textbf{0.9307}\\
			&&128 &0.8789 &0.8597 &0.9060 &0.9022 &0.8932 &0.8978 &0.9181 &\underline{0.9187} &0.9050 &0.8710 &\textbf{0.9305}\\
			\hline
	\end{tabular}}
\end{table}
\begin{figure}[h]
	\centering
	\subfigure[]{\includegraphics[width=1.78in]{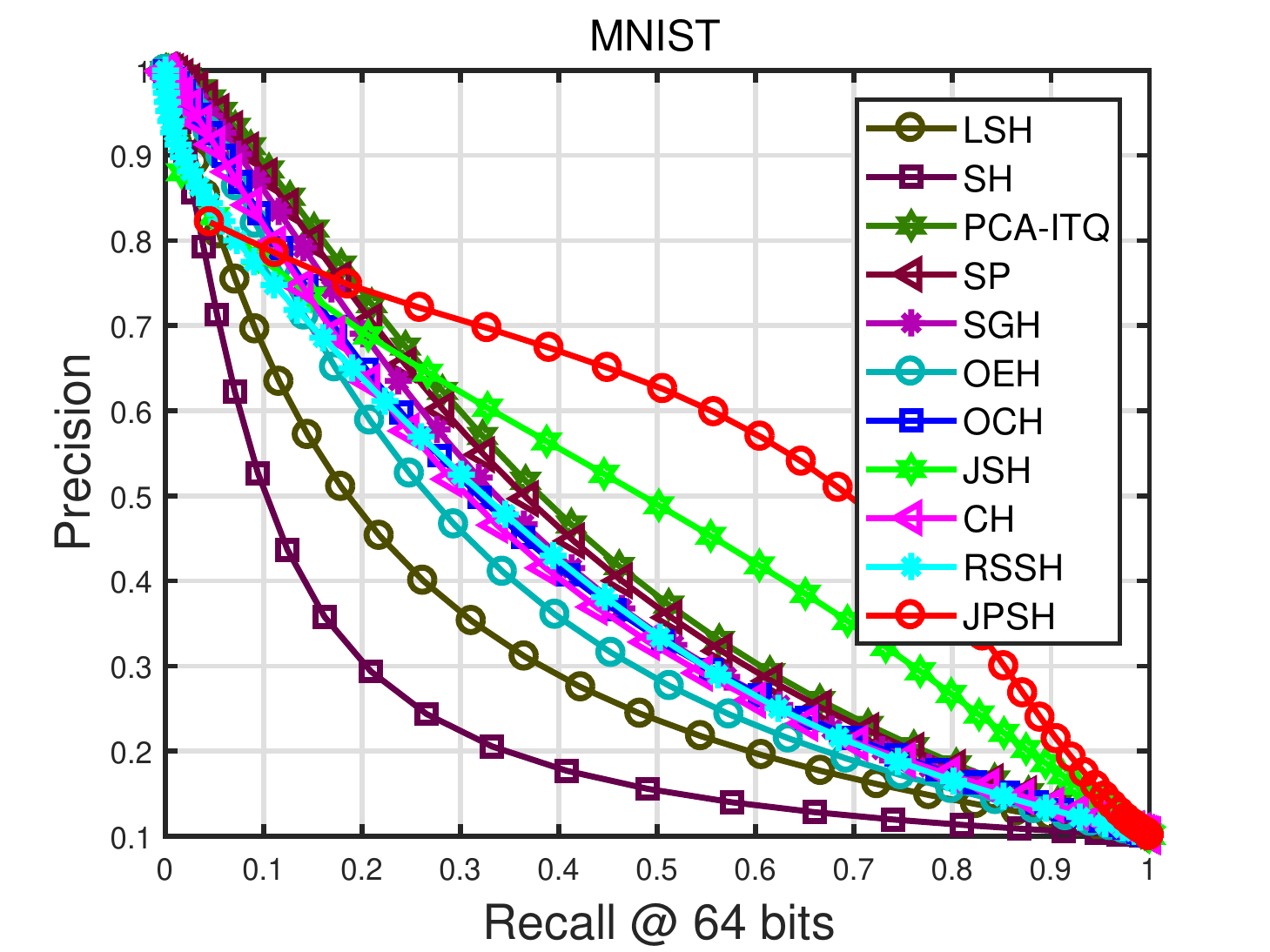}}
	\subfigure[]{\includegraphics[width=1.78in]{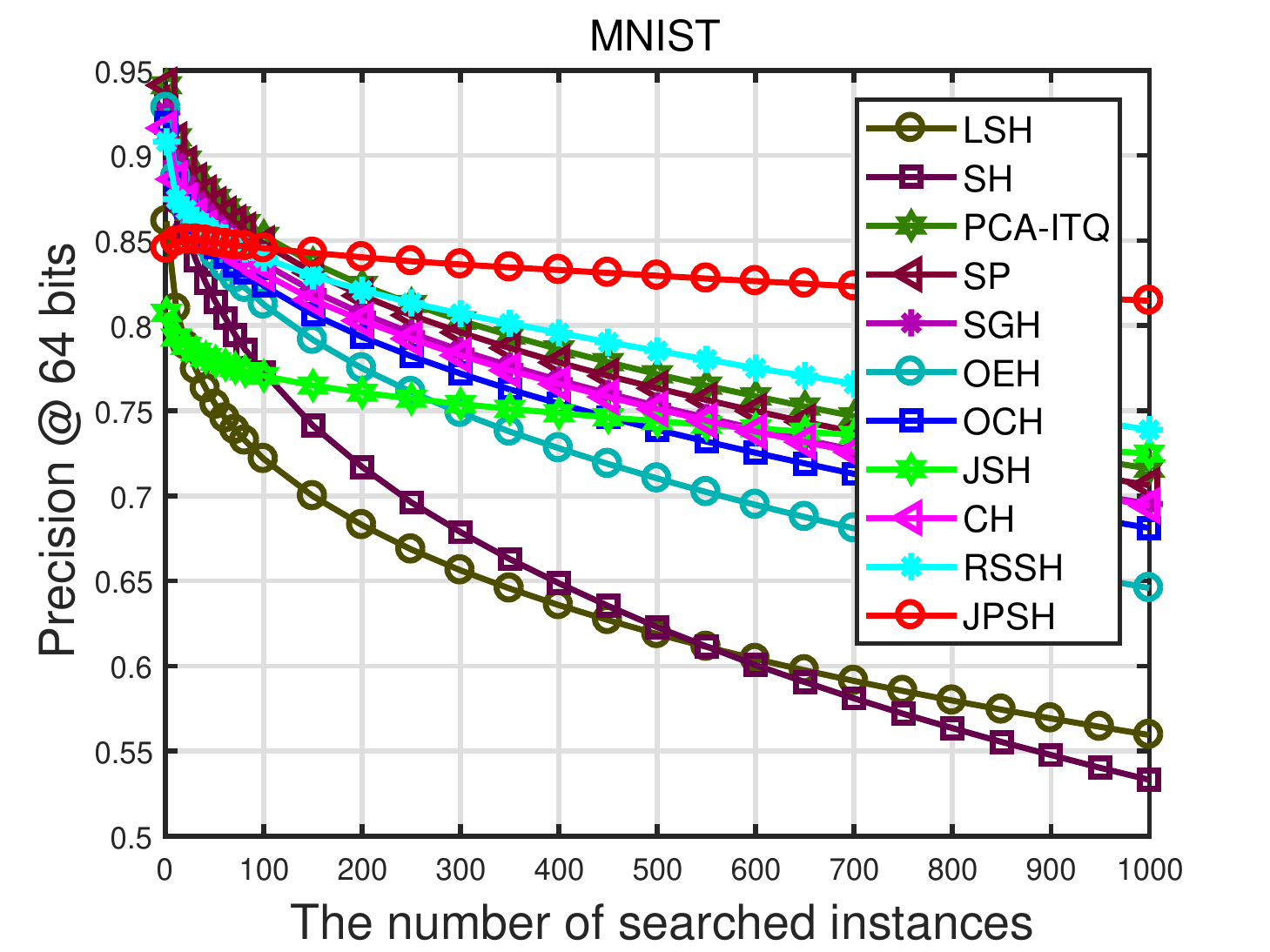}}
	\subfigure[]{\includegraphics[width=1.78in]{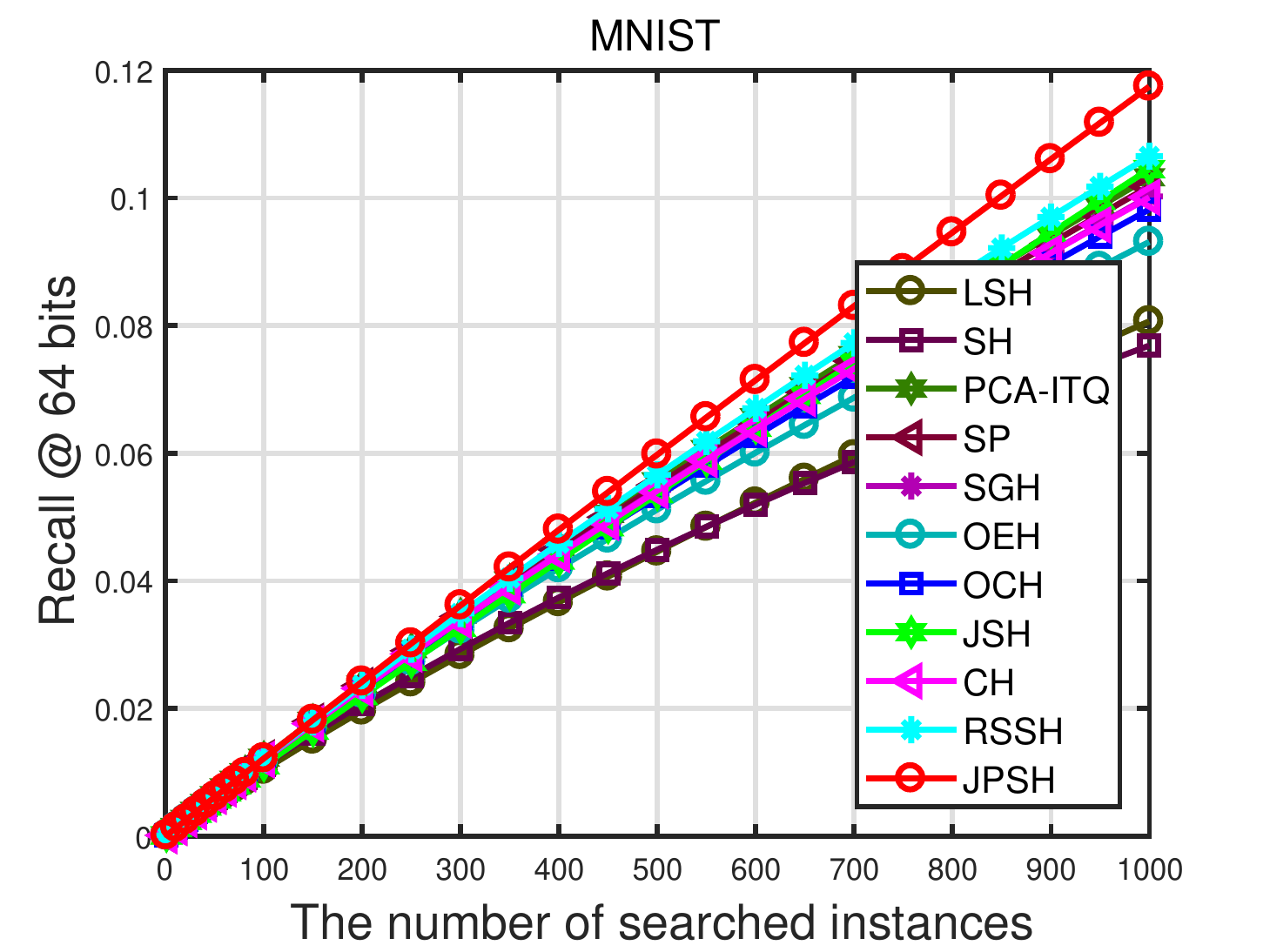}}
	\subfigure[]{\includegraphics[width=1.78in]{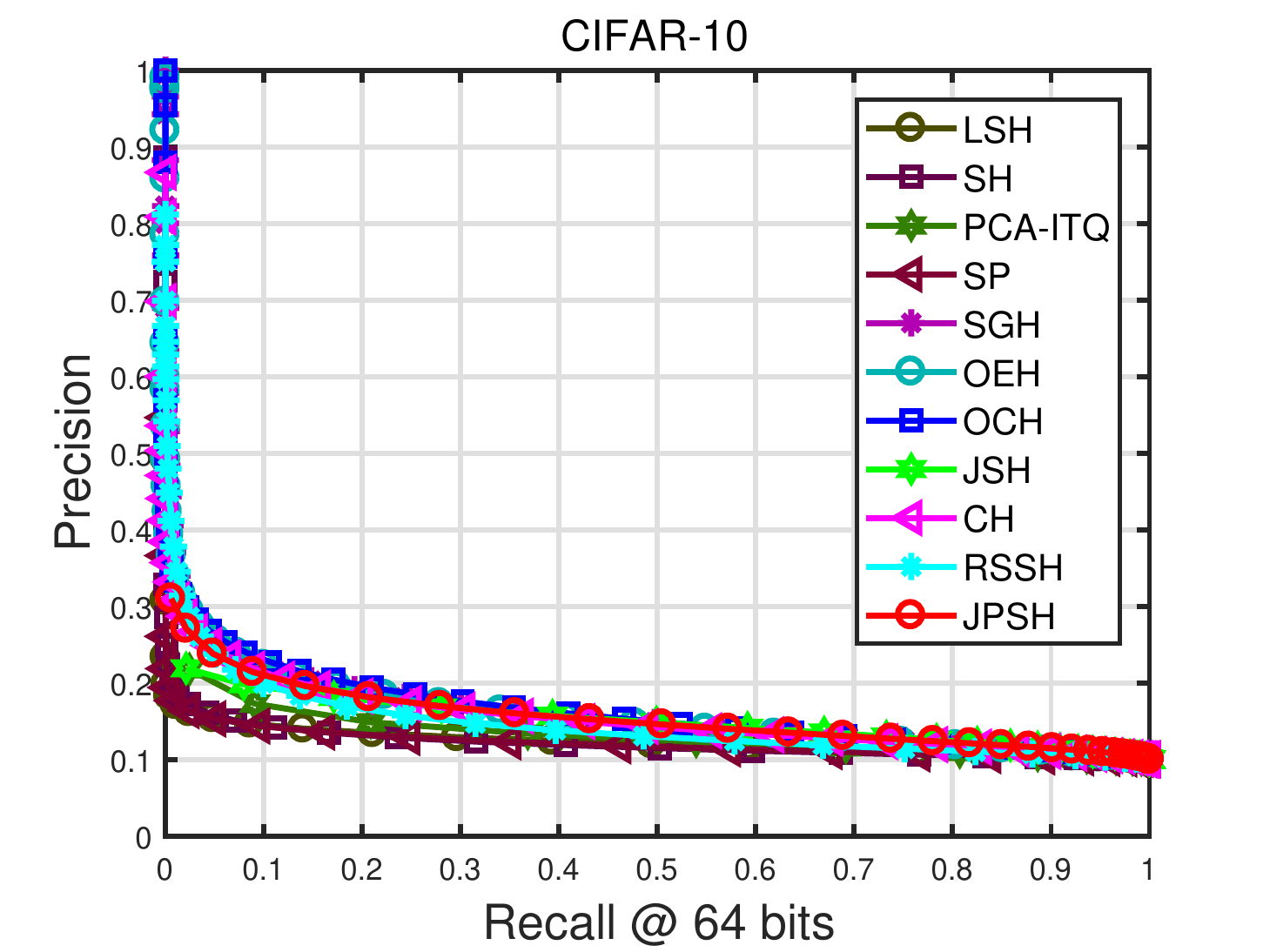}}
	\subfigure[]{\includegraphics[width=1.78in]{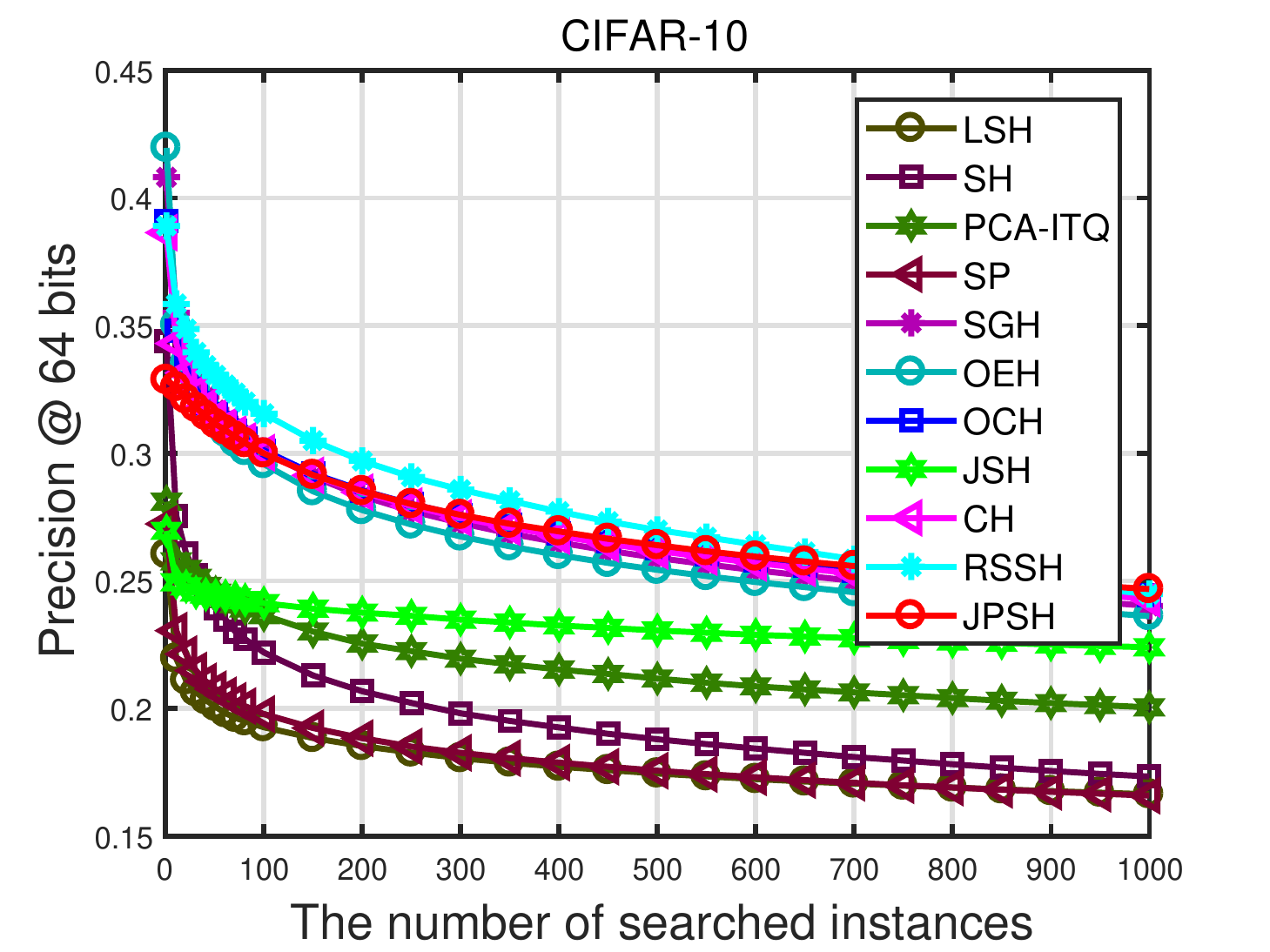}}
	\subfigure[]{\includegraphics[width=1.78in]{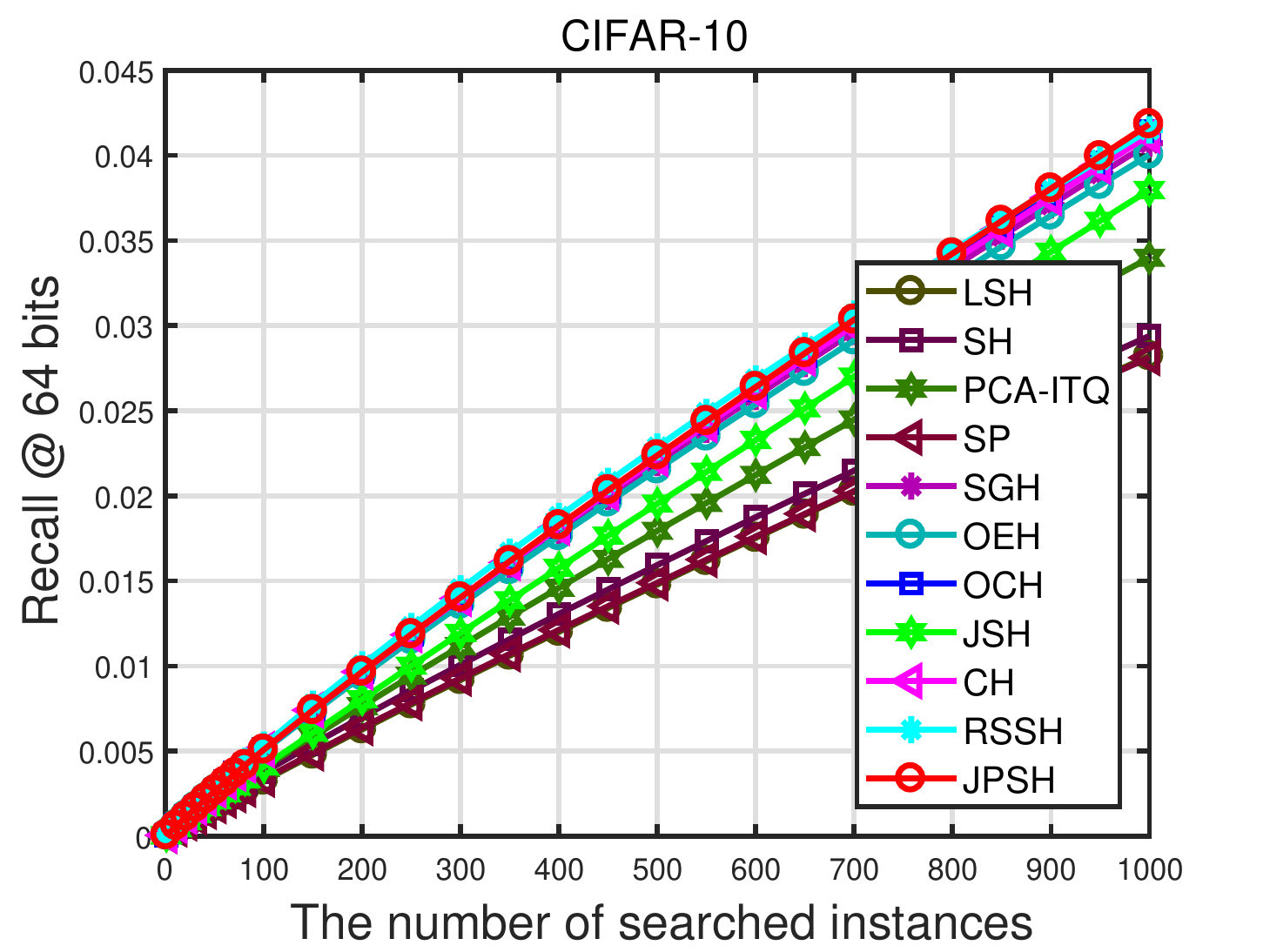}}
	\subfigure[]{\includegraphics[width=1.78in]{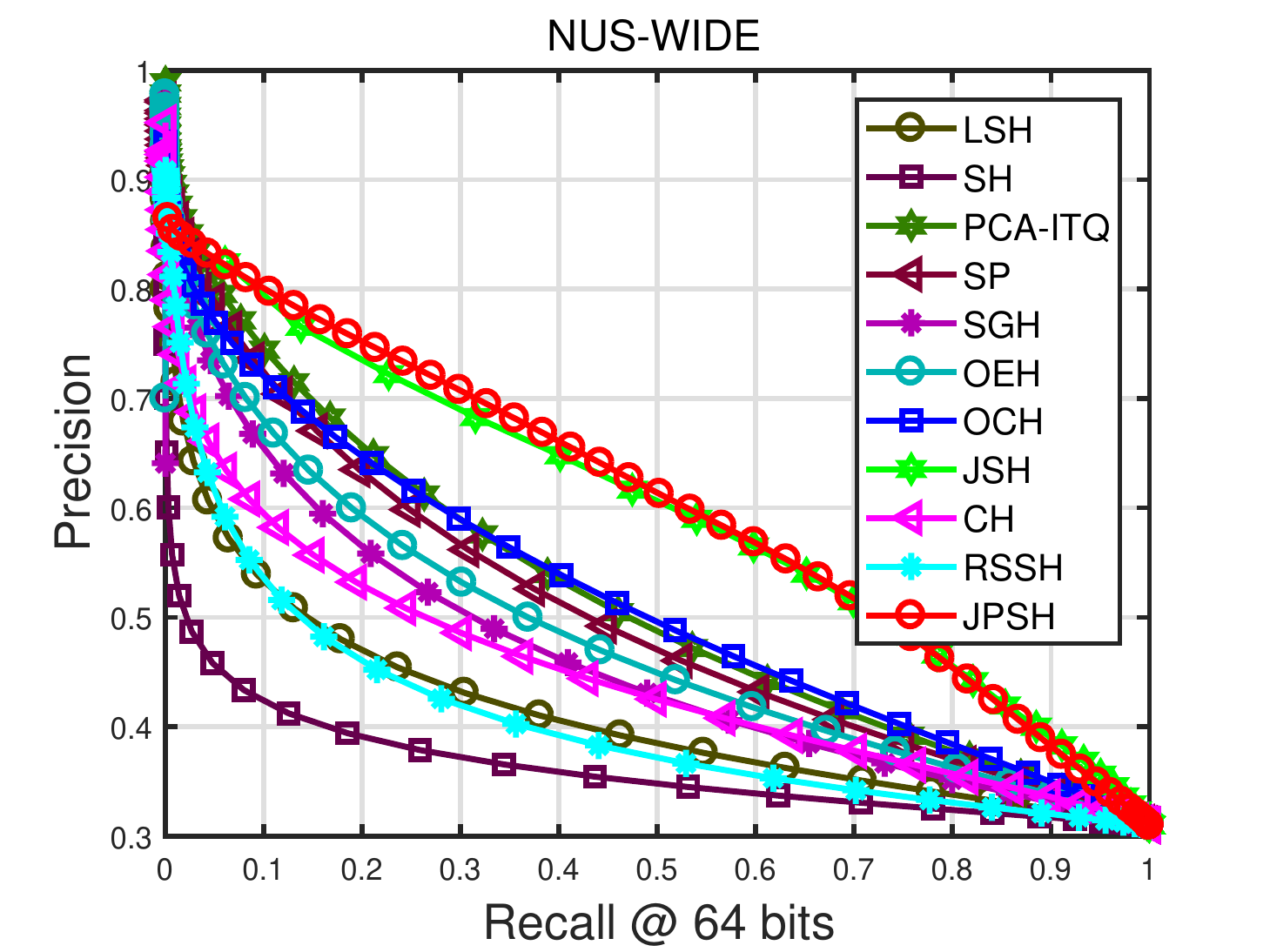}}
	\subfigure[]{\includegraphics[width=1.78in]{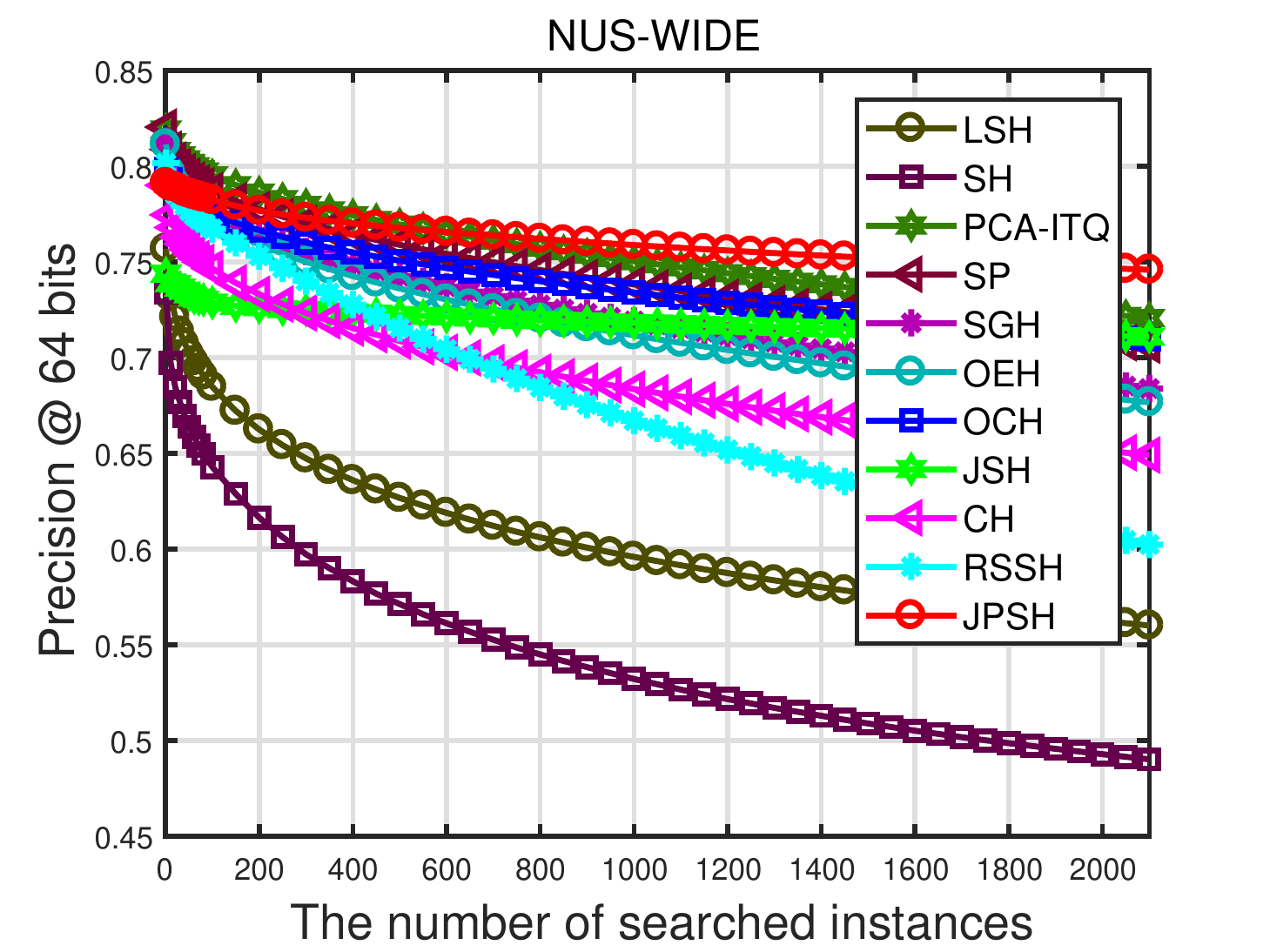}}
	\subfigure[]{\includegraphics[width=1.78in]{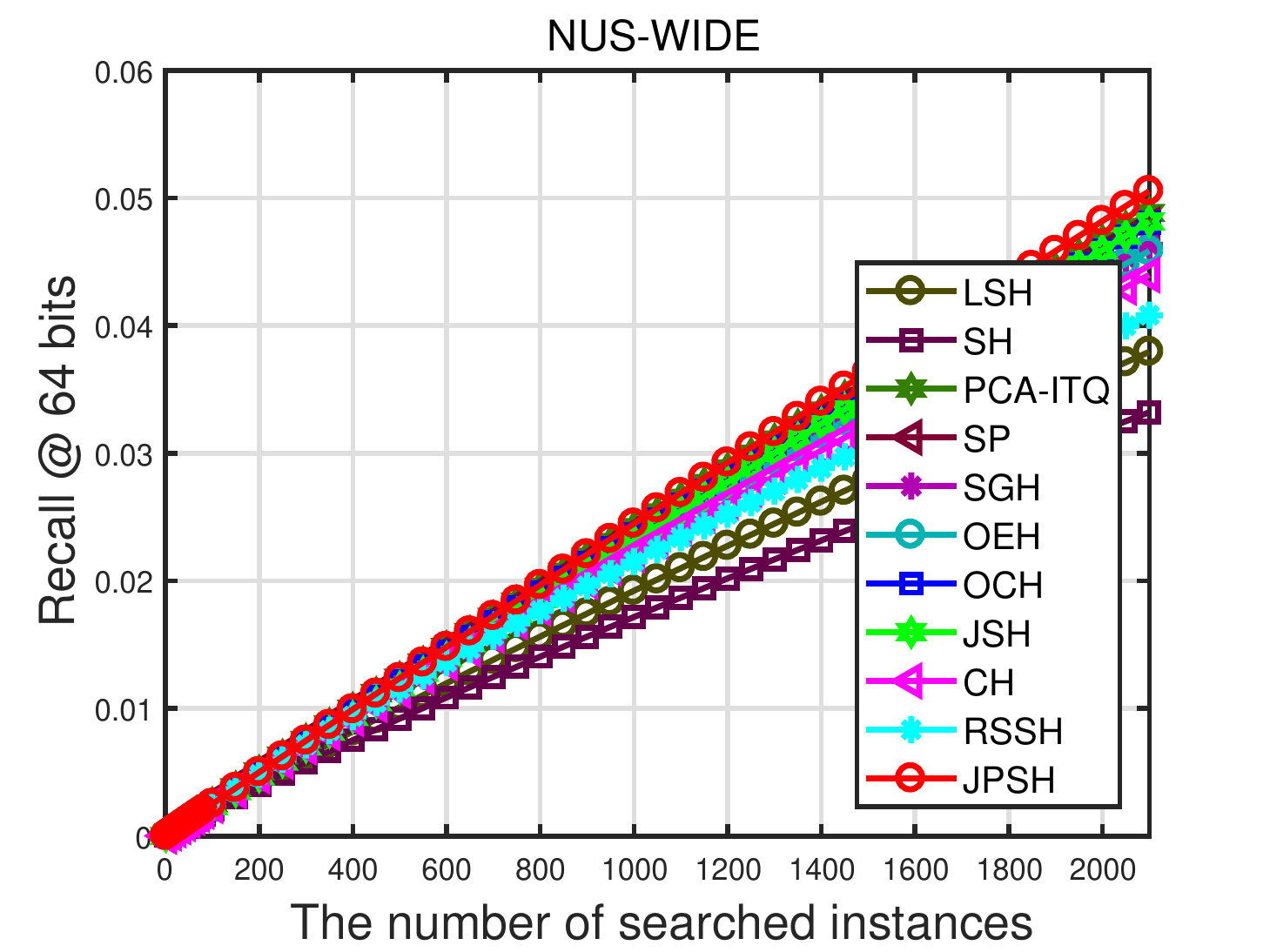}}
	\subfigure[]{\includegraphics[width=1.78in]{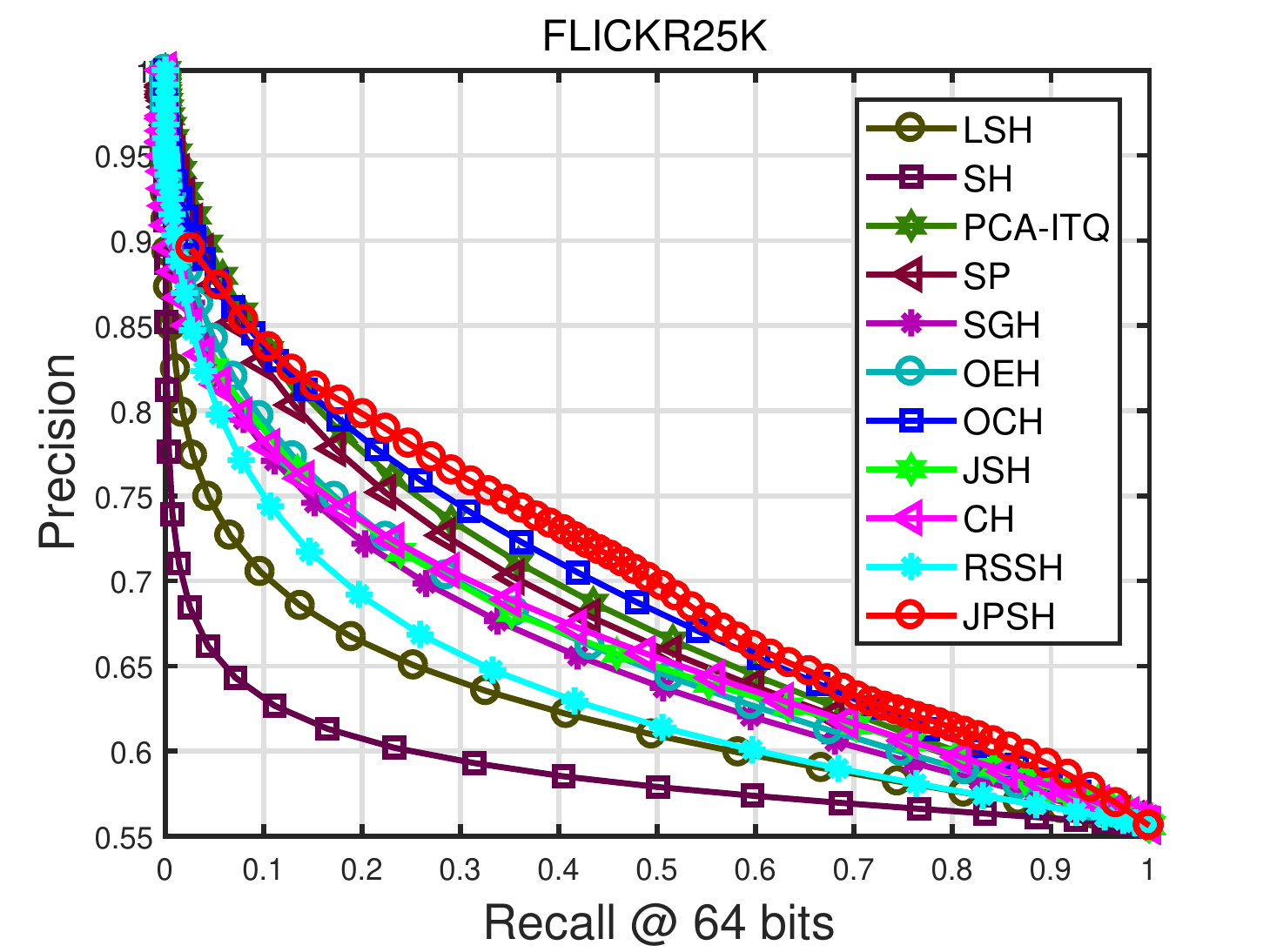}}
	\subfigure[]{\includegraphics[width=1.78in]{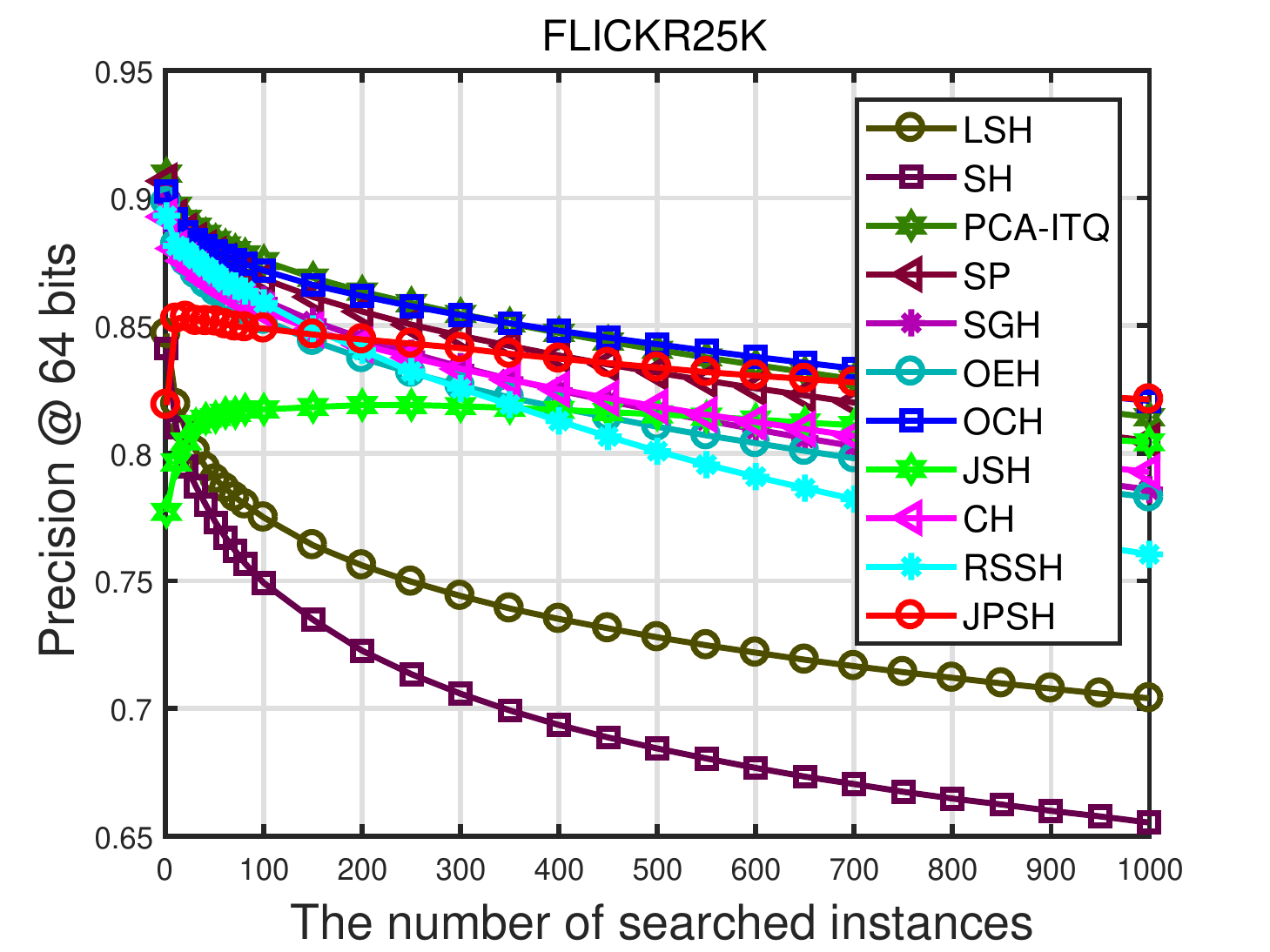}}
	\subfigure[]{\includegraphics[width=1.78in]{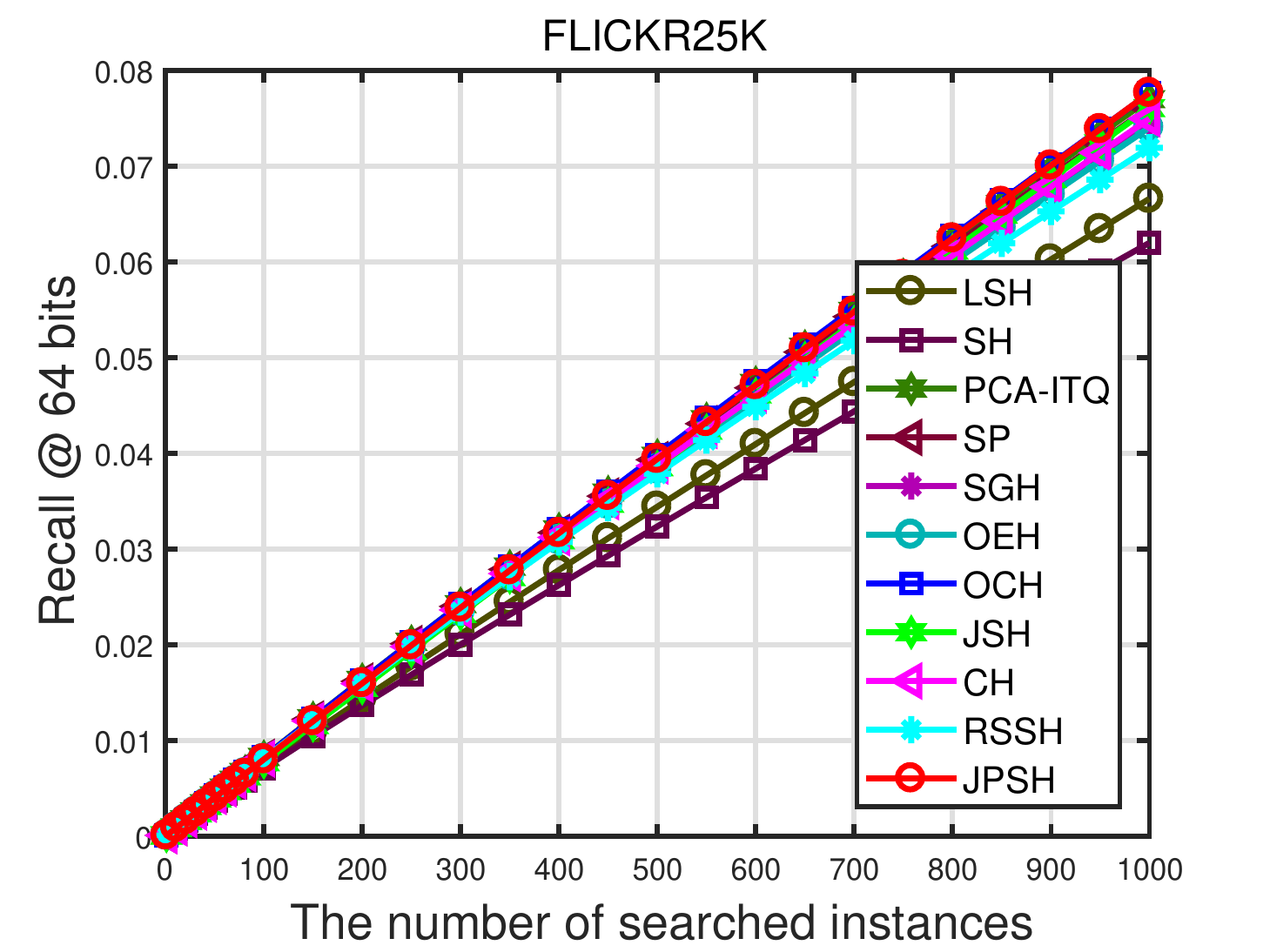}}
	\caption{Precision-Recall curve, Precision and Recall vs. the number of searched instances on the MNIST, CIFAR-10, NUS-WIDE and FLICKR25K datasets with 64 bits.}
	\label{fig:4}
\end{figure}

\subsection{Results and Discussions}
For the MNIST dataset, the first two rows in Table~\ref{tab:1} illustrate the mAP and Pre@100 of our proposed JPSH and the baselines with different hash bits. Figure~\ref{fig:4} (a, b, c) show the Precision-Recall curve, Precision@N and Recall@N curves on 64 bits, respectively. Going through these results, we have the following observations: (1) JPSH consistently outperforms all baselines in all cases on this dataset with large gaps, demonstrating its effectiveness on the binary representation in dealing with similarity search tasks. (2) Compared against the second-best method JSH, our proposed method improves $23.76\%$ and $8.63\%$ at least for mAP and Pre@100, respectively. This shows that JPSH outperforms the standard manifold-based hashing method JSH, and generates more discriminative binary codes by jointly learning semantic and pairwise similarities in the Hamming space. (3) RSSH shows better performance than manifold-based hashing methods in low hash bits but worse than JPSH. This indicates that semantic preservation helps to learn the discriminative information, especially that for our proposed method. (4) Figure~\ref{fig:4} (a, b, c) show that JPSH achieves consistent performance improvement in terms of Precision-Recall curve, Precision@N and Recall curves on 64 bits, indicating the effectiveness of our JPSH.

For the CIFAR-10 dataset, mAP and Pre@100 are reported in the third and fourth rows of Table~\ref{tab:1}, and Figure~\ref{fig:4} (d, e, f) show the Precision-Recall curve, Precision@N and Recall@N curves based on 64 bits, respectively. From these results, we have the observations similar to those of MNIST: (1) Compared with all baselines, JPSH achieves the best results of 16 to 128 bits in terms of mAP and Pre@100, demonstrating that JPSH can learn discriminative binary codes by maintaining semantic structures in different personalized subspaces. (2) Of 8 bits, PCA-ITQ achieves the best performance. However, the performance of JPSH and PCA-ITQ is very close, whilst JPSH is competitive to the other methods. (3) In Figure~\ref{fig:4} (d, e, f), JPSH matches the best baseline in terms of Precision-Recall curve and Recall@N curve, but little low than RSSH on Precision@N. It is worth noting that JPSH ourperforms JSH in terms of Precision@N and Recall@N curves. This indicates that PSH can help to map instances within the same cluster to the same Hamming space in a certain extent.

For the NUS-WIDE dataset, we report the results of mAP and Pre@100 in the fifth and sixth rows of Table~\ref{tab:1}. The Precision-Recall, Precision@N, and Recall@N curves are shown in Figure~\ref{fig:4} (g, h, i). Regarding those results, we have similar observations to the above cases: (1) JPSH also outperforms all baselines in all cases in terms of mAP on this real-world dataset. It significantly surpasses all baselines of low hash bits (8 to 32 bits) in terms of Pre@100, and little lower than JSH of high hash bits (64 to 128 bits). As a whole, these results demonstrate the effectiveness and robustness of the semantic- and pairwise-preserving scheme used in JPSH. (2) From Figure~\ref{fig:4} (g, h, i), we witness that JPSH has a consistent improvement compared with all baselines. This justifies that our proposed JPSH discovers more discriminative structural information in binary representation learning.

For the FLICKR25K dataset, the last two rows of Table~\ref{tab:1} show results of mAP and Pre@100, and curves of Precision-Recall, Precision@N and Recall@N are shown in Figure~\ref{fig:4} (j, k, l). Again, we have observations similar to the above three datasets: (1) JPSH obtains the best results compared with all baselines in terms of mAP and Pre@100, demonstrating the effectiveness and robustness of our proposed JPSH in the binary representation learning process. (2) Figure~\ref{fig:4} (j, k, l) show that our proposed JPSH outperforms all compared methods in terms of Precision-Recall and Recall@N curves, slightly lower than some compared methods in terms of Precision@N curve. Furthermore, JPSH is always better than JSH in Figure~\ref{fig:4} (k). Therefore, our JPSH can learn better representative binary codes for similarity search.
\begin{figure}[h]
	\centering
	\subfigure[]{\includegraphics[width=1.78in]{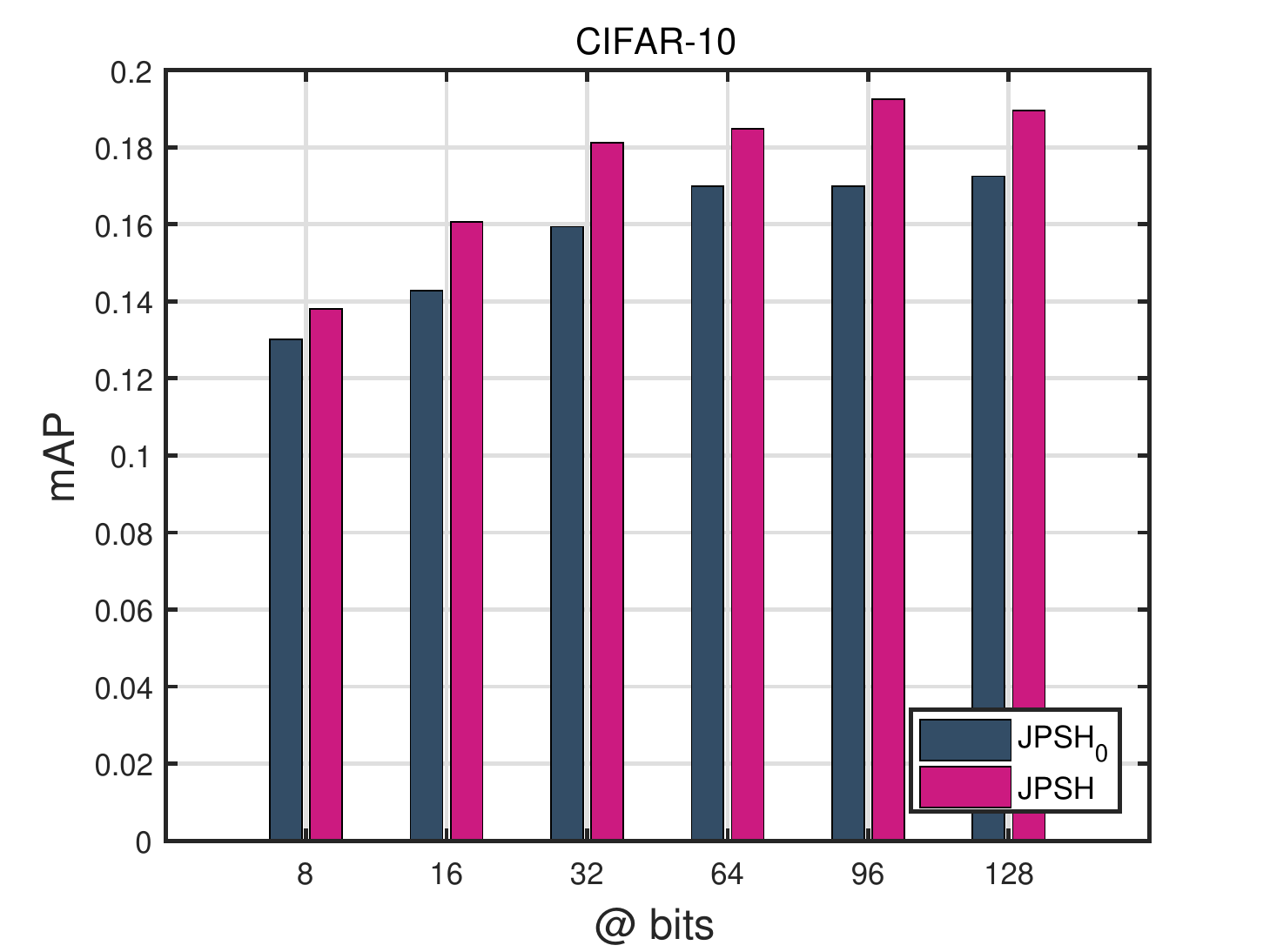}}
	\subfigure[]{\includegraphics[width=1.78in]{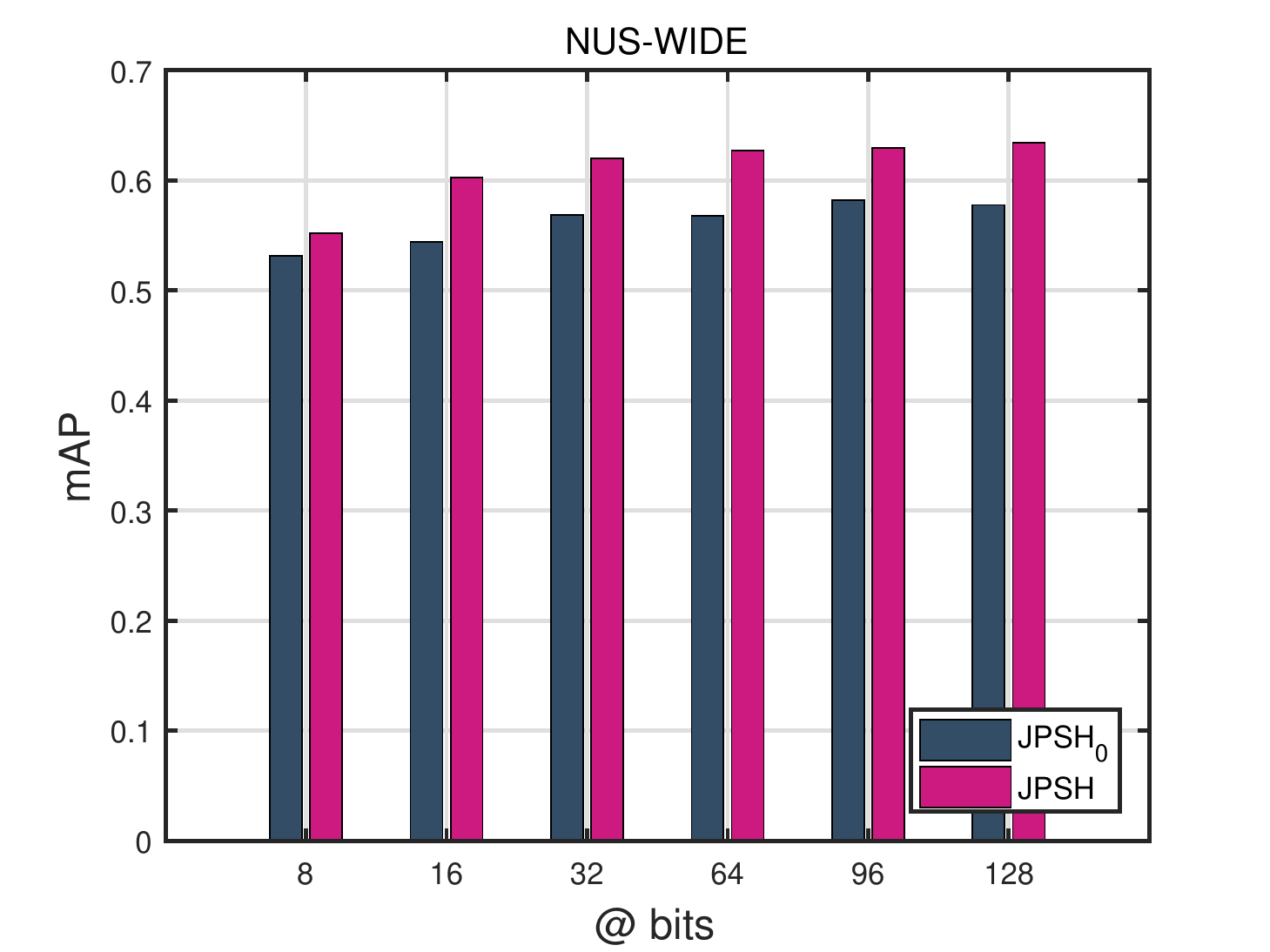}}
	\subfigure[]{\includegraphics[width=1.78in]{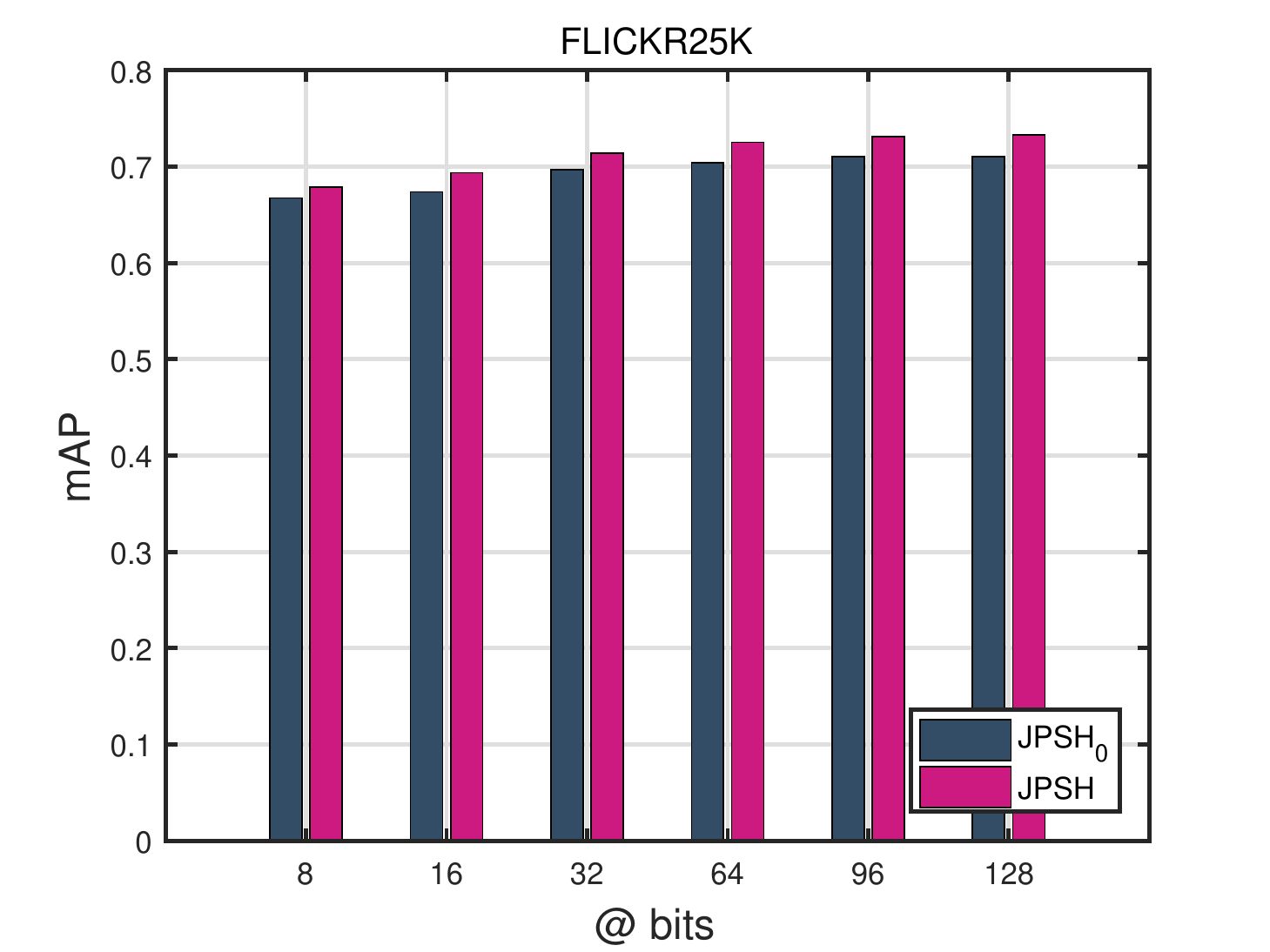}}
	\caption{Compared results of mAP between JPSH$_0$ and JPSH on (a) CIFAR-10, (b) NUS-WIDE and (c) FLICKR25K datasets.}
	\label{fig:5}
\end{figure}
\begin{figure}[h]
	\centering
	\subfigure[]{\includegraphics[width=1.78in]{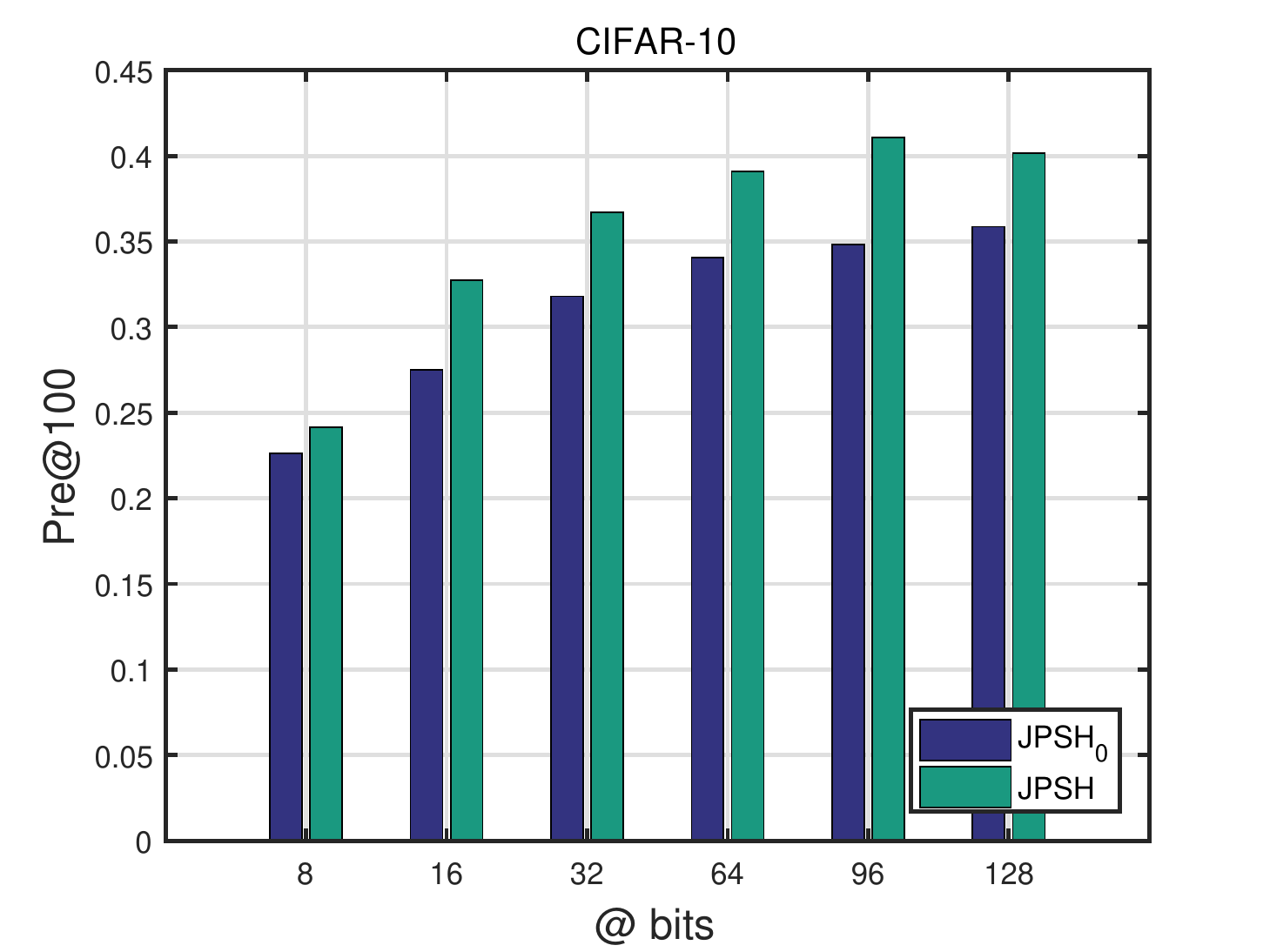}}
	\subfigure[]{\includegraphics[width=1.78in]{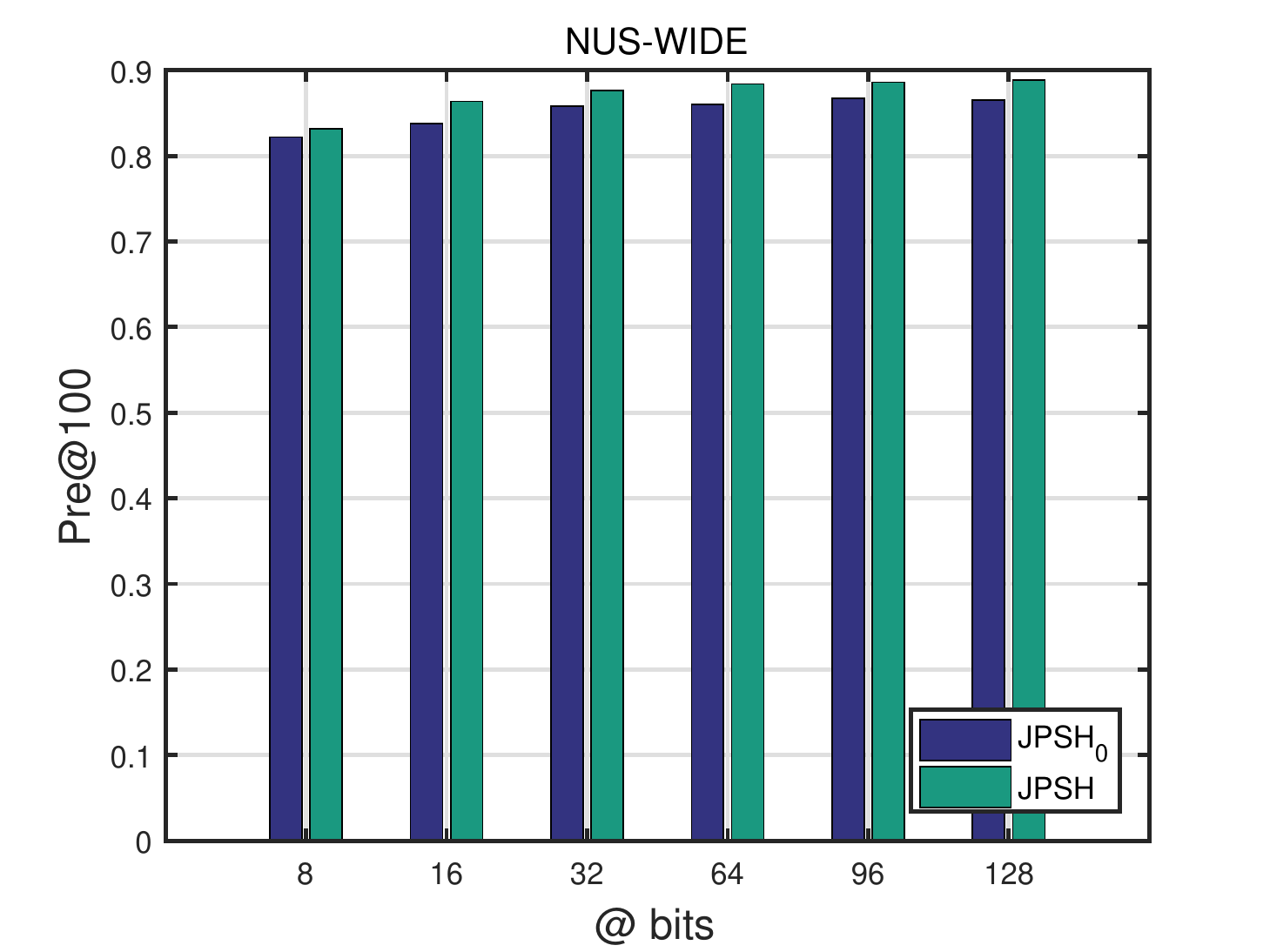}}
	\subfigure[]{\includegraphics[width=1.78in]{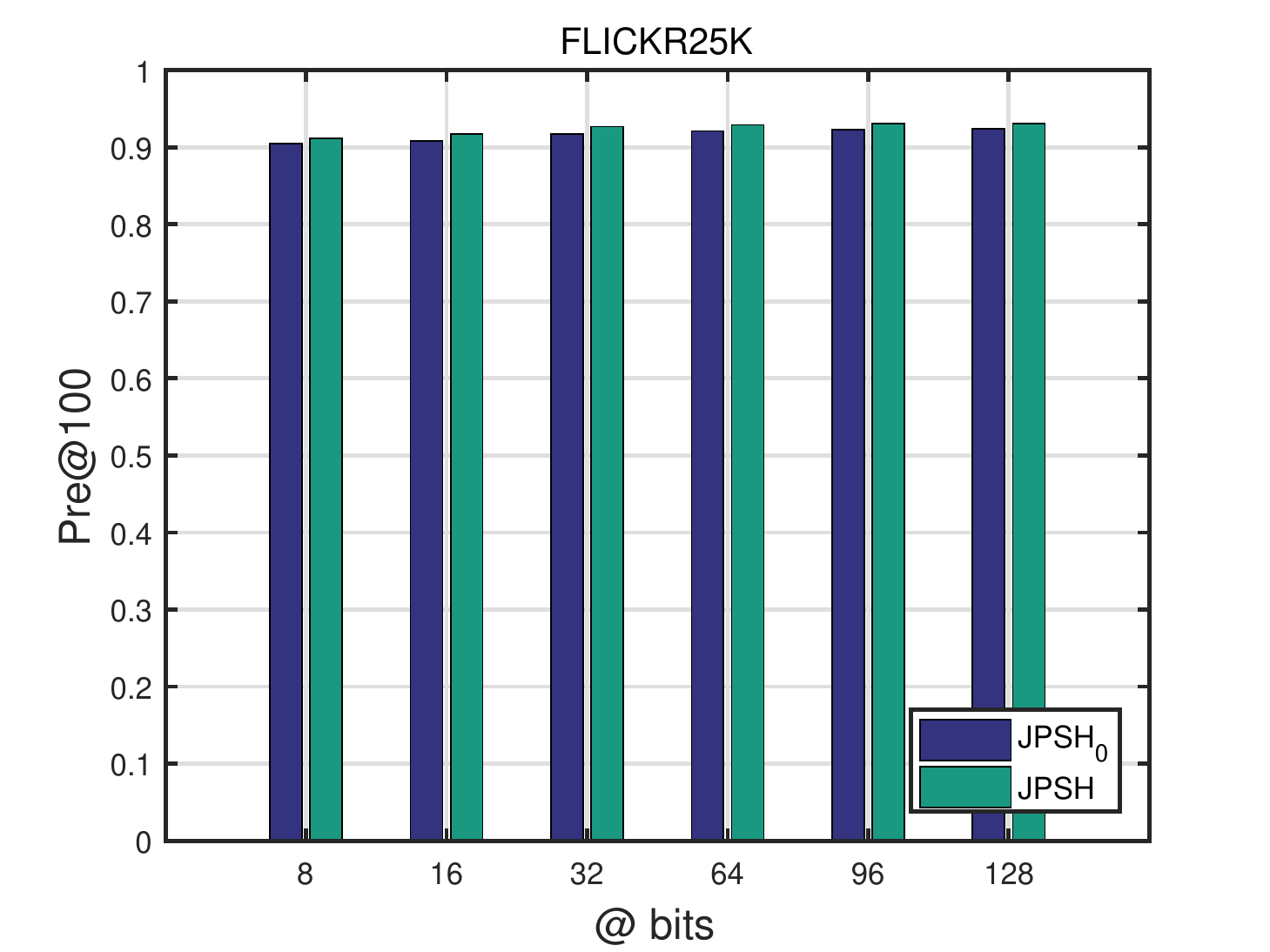}}
	\caption{Compared results of Pre@100 between JPSH$_0$ and JPSH on (a) CIFAR-10, (b) NUS-WIDE and (c) FLICKR25K datasets.}
	\label{fig:6}
\end{figure}

\subsection{Semantic Analysis}
In JPSH, we use pseudo labels to distinguish different clusters, and then map instances within the same cluster to the same Hamming space. In order to verify that our JPSH indeed maintains a certain semantic similarity, the training set of PSH is replaced by random samples called JPSH$_0$. The results of mAP and Pre@100 between JPSH$_0$ and JPSH are shown in Fig~\ref{fig:5} and Fig~\ref{fig:6}, respectively.  
From those figures, we can observe that the mAP and Pre@100 of JPSH$_0$ are obviously below that of JPSH, particularly in CIFAR-10 and NUS-WIDE datasets. These results indicate that K-means is good for extracting discriminative information with exploring the semantic structures. Moreover, our proposed JPSH successfully retains those structures into the Hamming space, and JPSH$_0$ can not be.

\begin{figure}[h]
	\centering
	\subfigure[]{\includegraphics[width=1.78in]{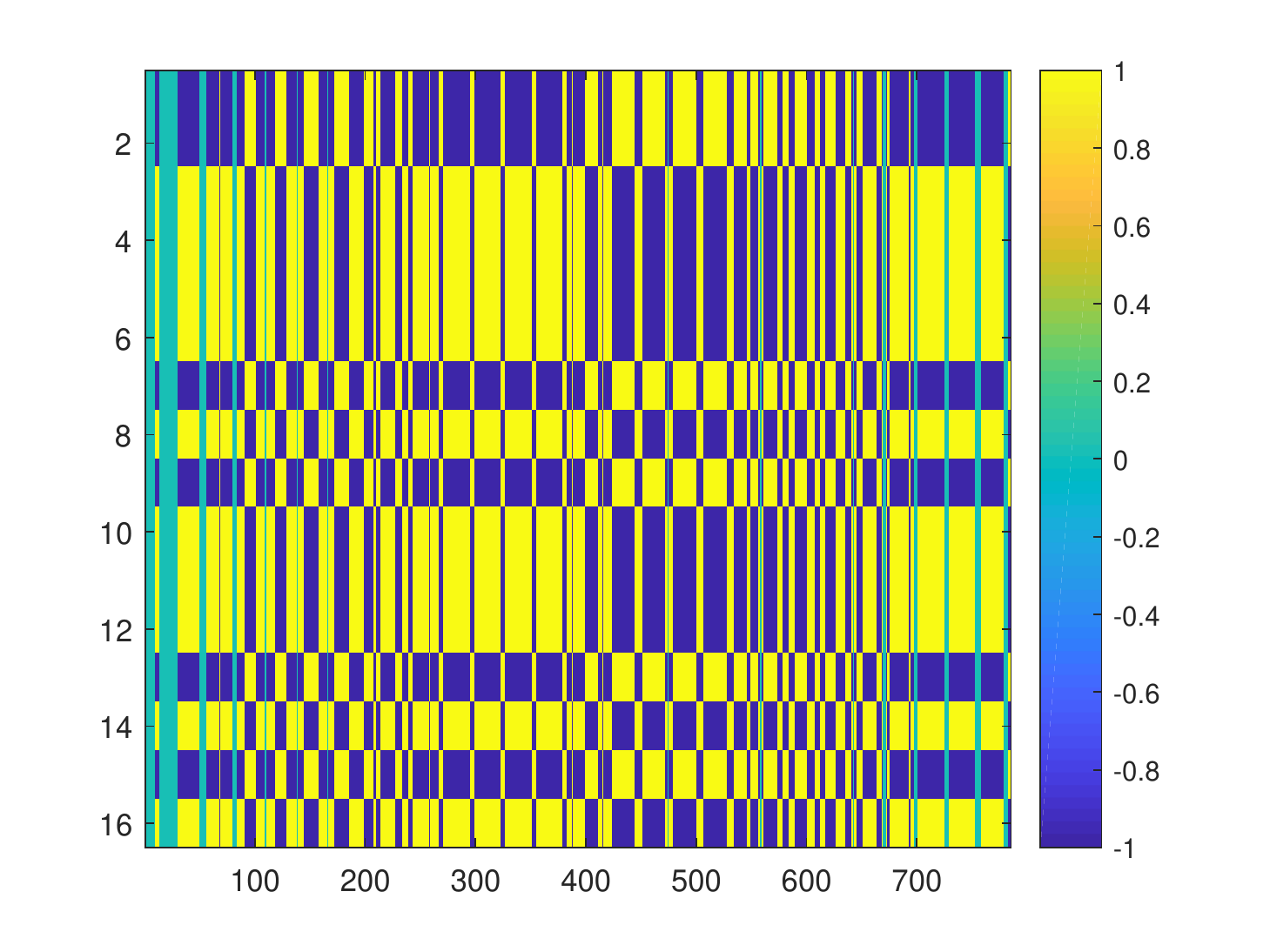}}
	\subfigure[]{\includegraphics[width=1.78in]{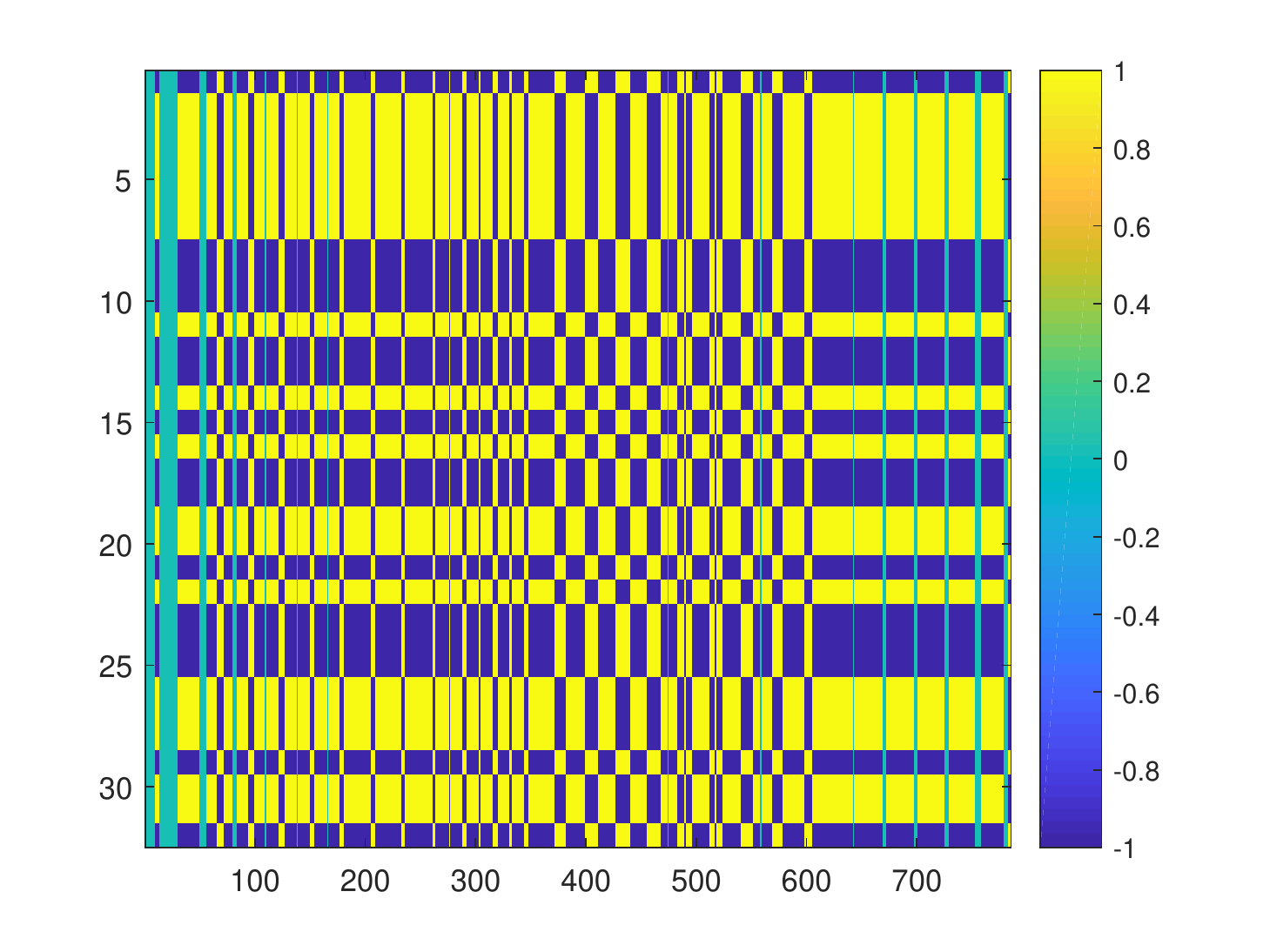}}
	\subfigure[]{\includegraphics[width=1.78in]{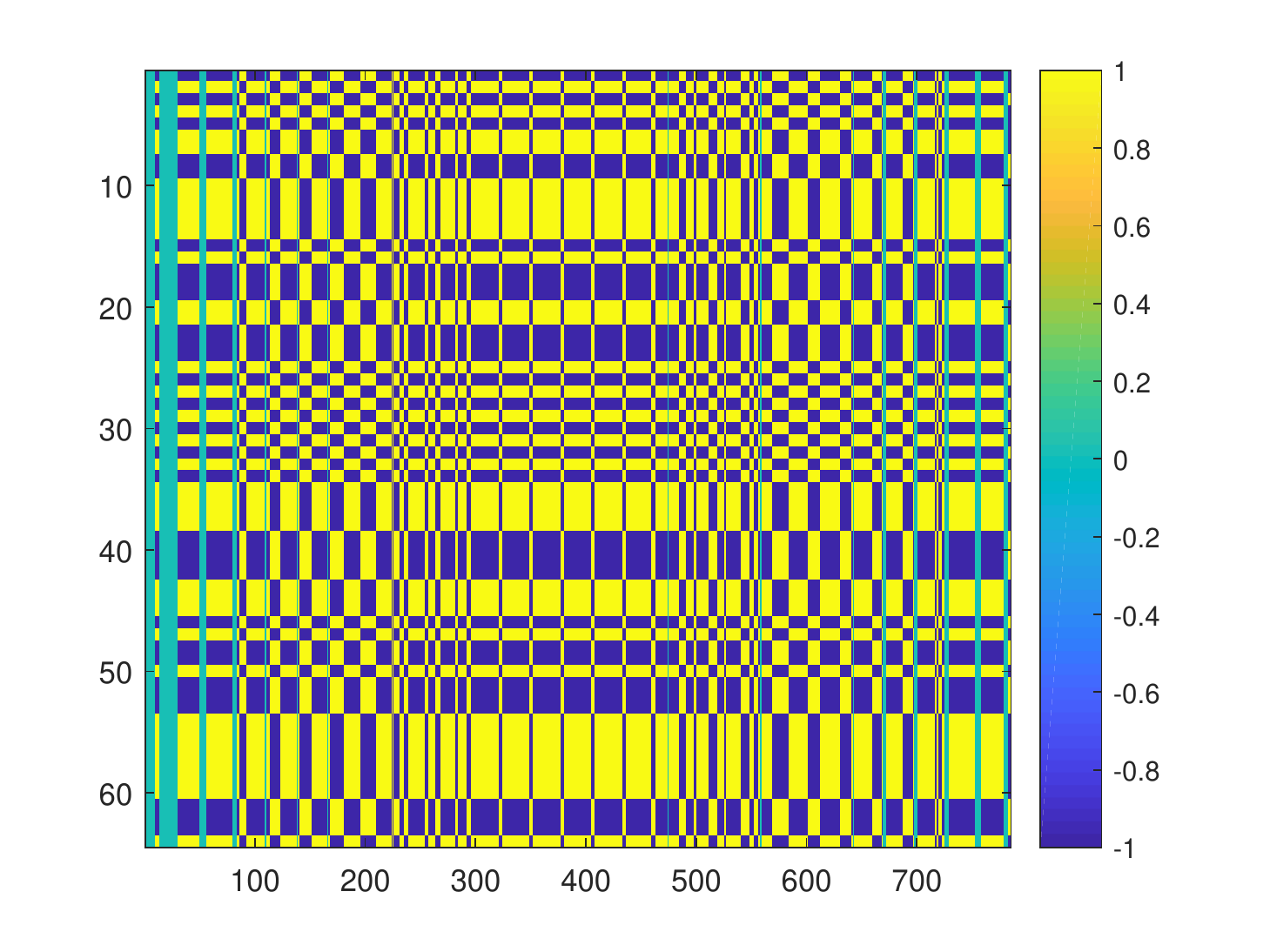}}
	\caption{Visualization of personalized weight matrix $\boldsymbol{\rm P}$ processed by ${\rm sign}$ function, (a) $16\times 784$ matrix, (b) $32\times 784$ matrix, and (c) $64\times 784$ matrix on MNIST dataset. The green column indicates that the corresponding features are zero and ignored during the projection operation.}
	\label{fig:7}
\end{figure}

\subsection{Sparsity Analysis}
As shown in Figure~\ref{fig:7}, to study the sparsity of JPSH, we depict the visualization of personalized weight matrix $\boldsymbol{\rm P}$ processed by ${\rm sign}$ function with size $16\times784$, $32\times784$ and $64\times784$. It can be seen that each matrix contains some green columns that are equal to zero due to the sparsity constraint of $\|\boldsymbol{\rm P}_{j}\|_{2,1}$. It means that those corresponding features are ignored in a projection operation. Thus, we can eliminate some redundant features for binary representation learning. For example, the first column of matrices in Figure~\ref{fig:7} (a), (b) and (c) are all the zero vector, which means that the first feature of data is malfunction to distinguish the differences between data. Another benefit of sparsity lies in that it can increase the interpretability of the JPSH model. We can know that which features provide the discriminative information, while those features do not. Therefore, the sparsity of $\|\boldsymbol{\rm P}_{j}\|_{2,1}$ works in our JPSH model.

\begin{figure}[h]
	\centering
	\subfigure[]{\includegraphics[width=1.78in]{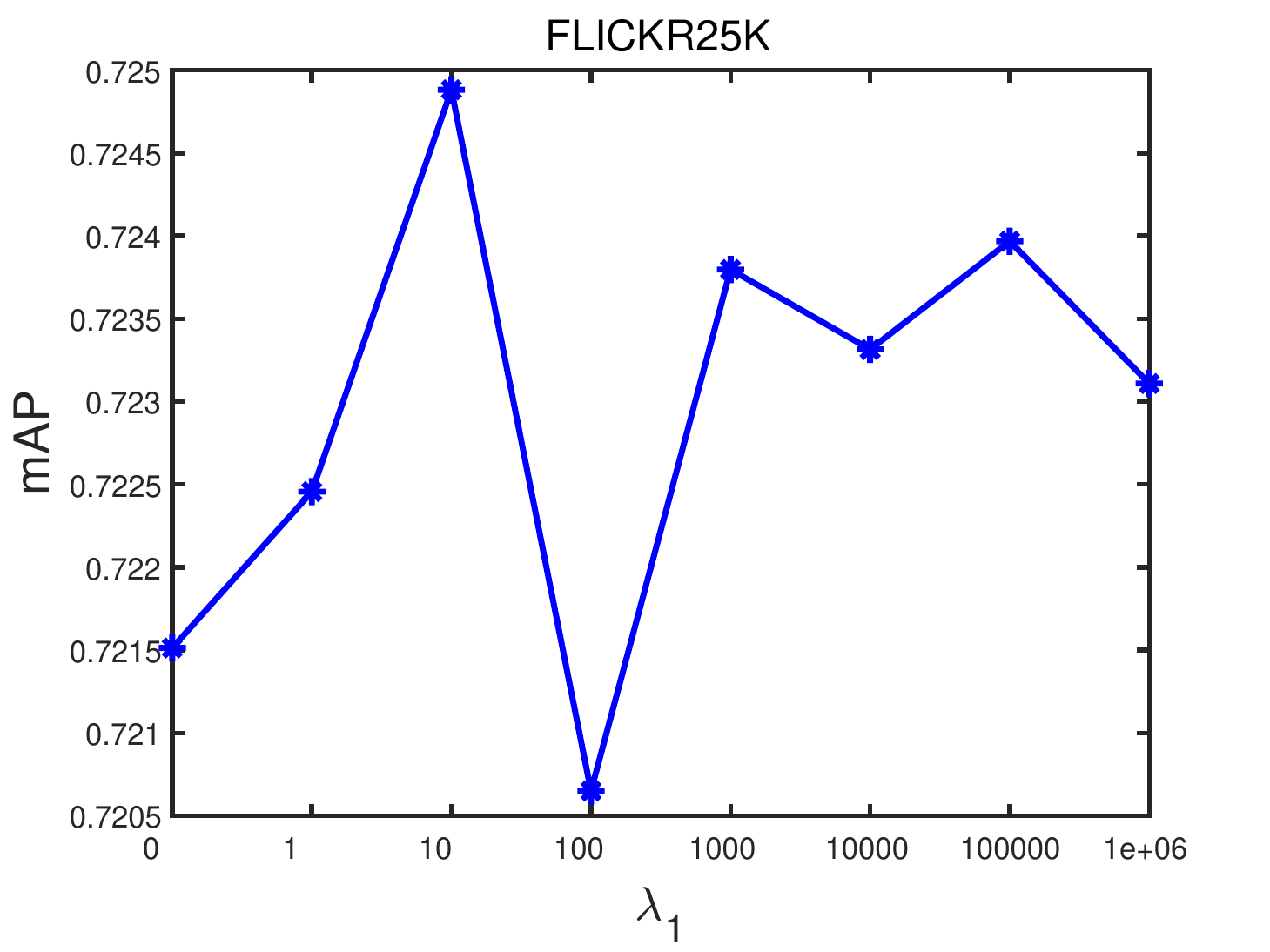}}
	\subfigure[]{\includegraphics[width=1.78in]{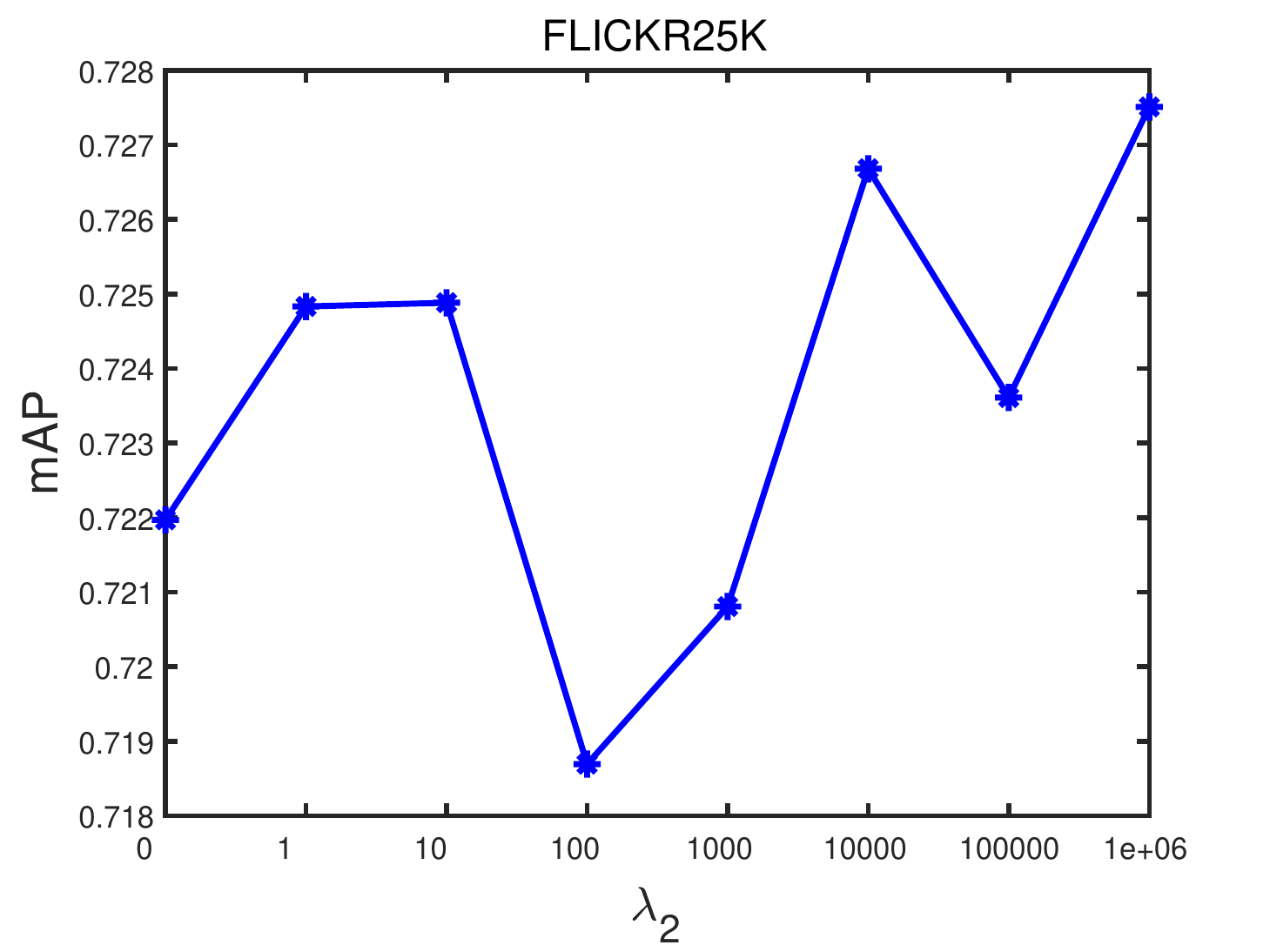}}
	\subfigure[]{\includegraphics[width=1.78in]{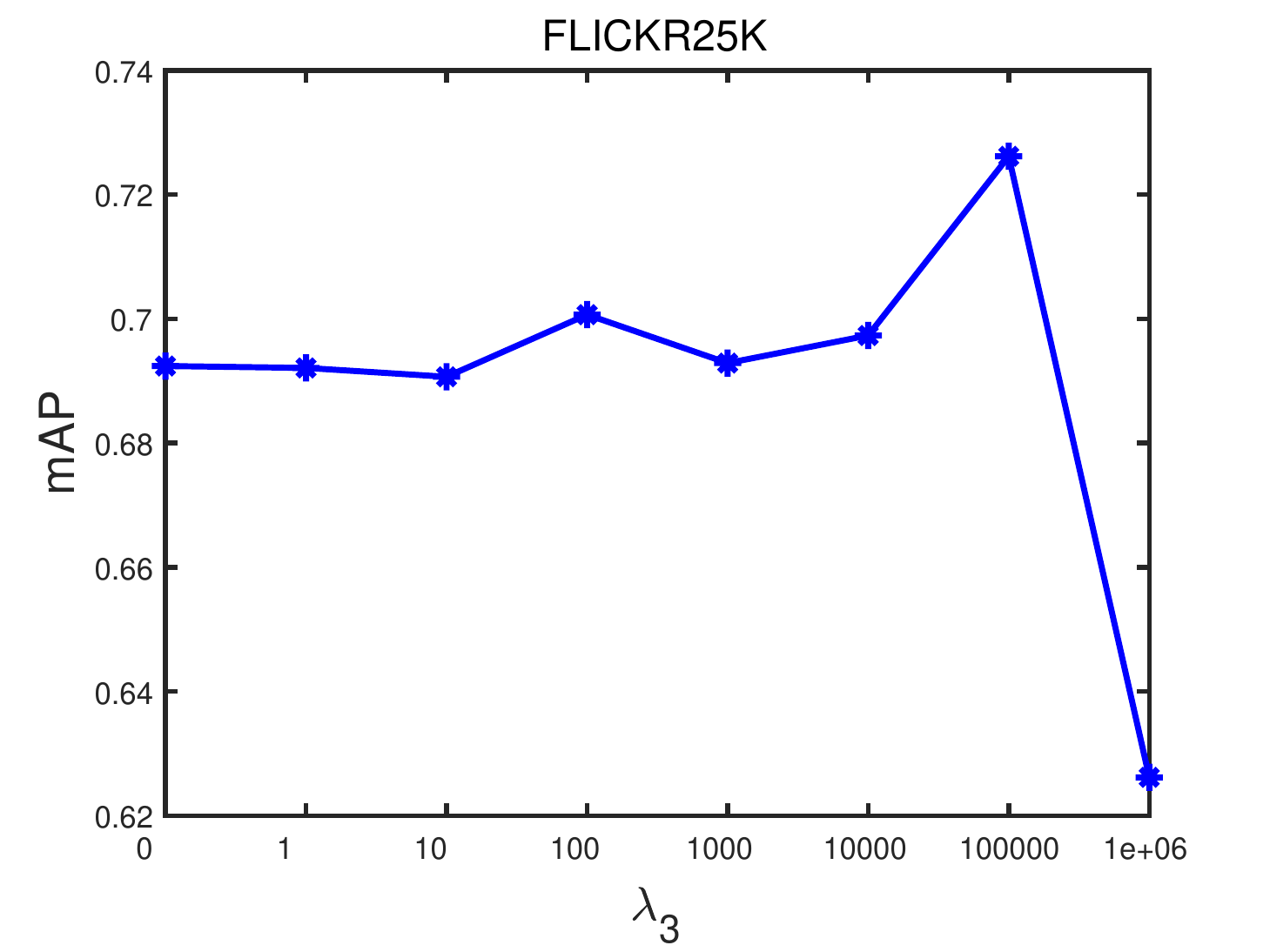}}
	\subfigure[]{\includegraphics[width=1.78in]{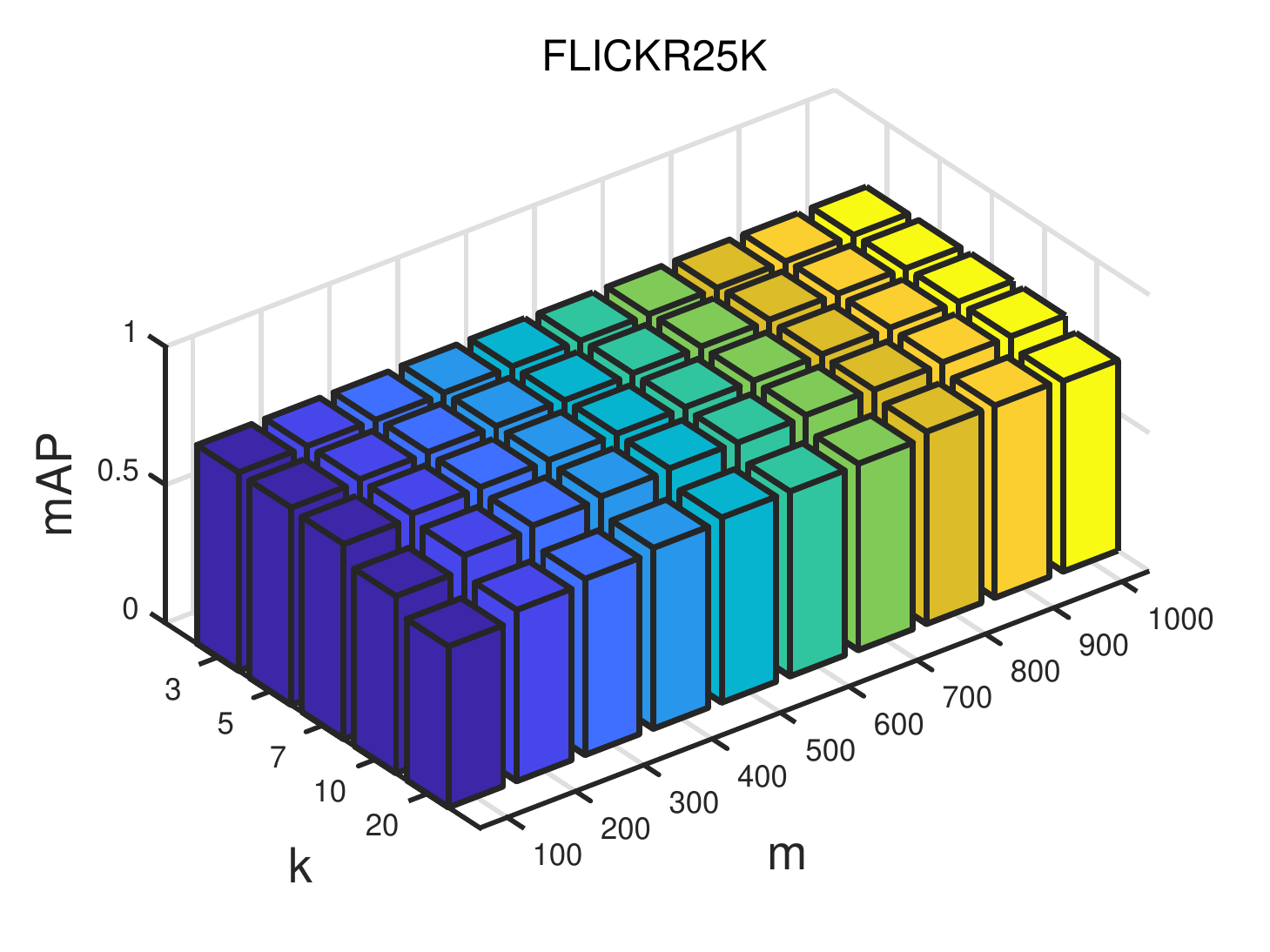}}
	\subfigure[]{\includegraphics[width=1.78in]{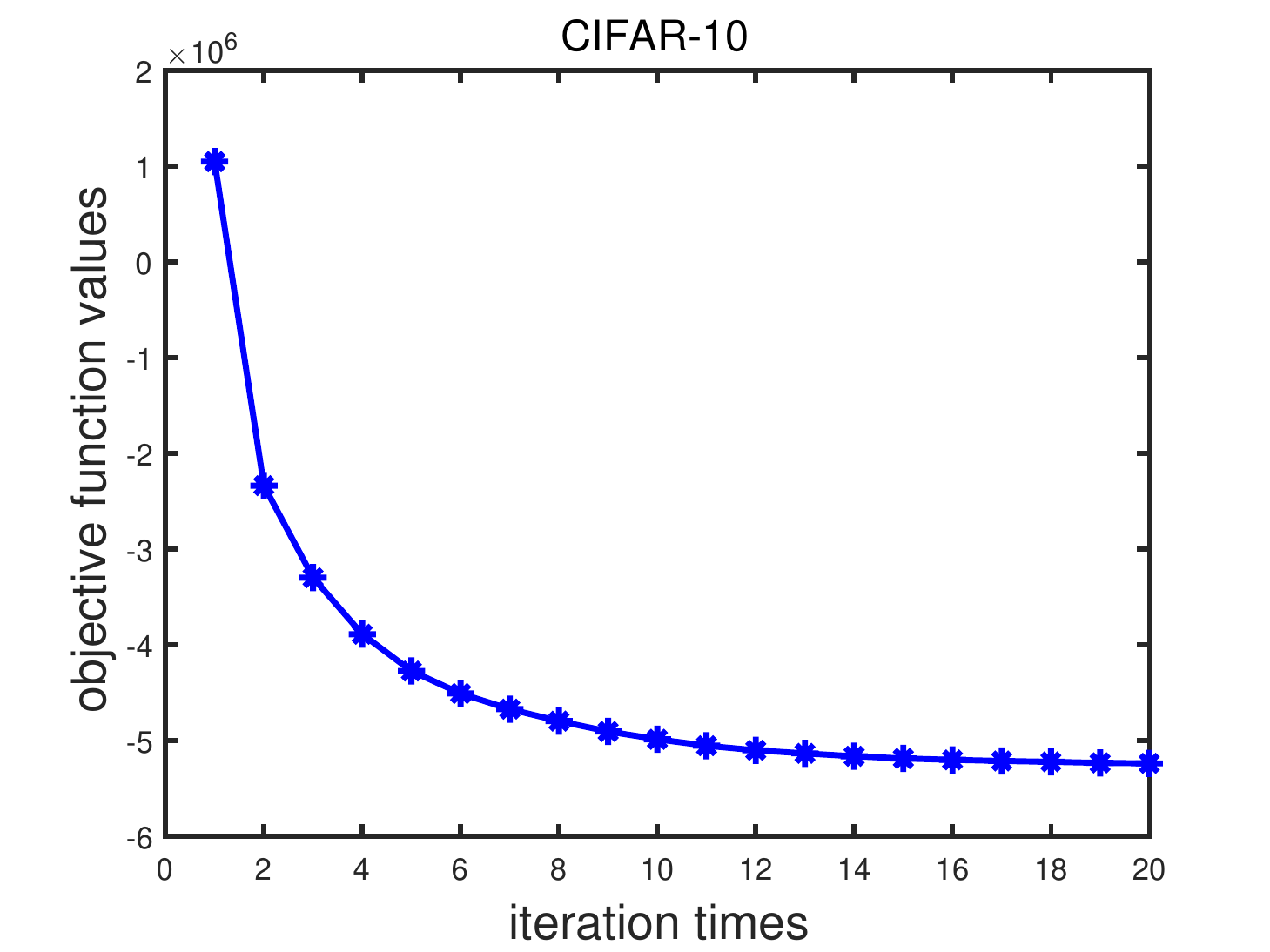}}
	\subfigure[]{\includegraphics[width=1.78in]{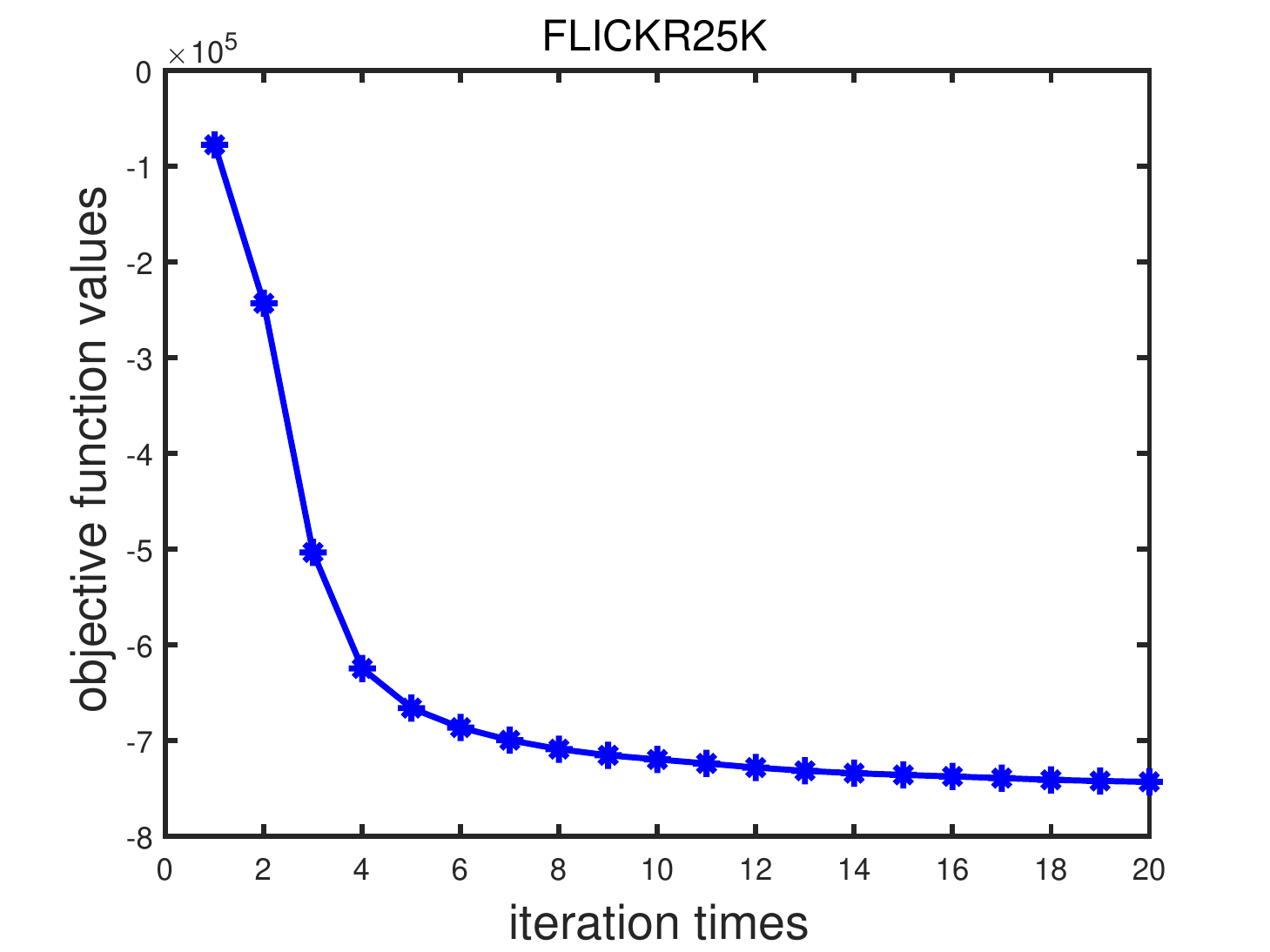}}
	\caption{Sensitivity analysis of $\lambda_1$, $\lambda_2$, $\lambda_3$, $k$ and $m$ in terms of mAP on the FLICKR25K dataset with 64 bits, and convergence curves of the JPSH on CIFAR-10 and FLICKR25K datasets with 32 bits.}
	\label{fig:8}
\end{figure}

\subsection{Sensitivity Analysis}
We discuss the sensitivity of our JPSH over parameters $\lambda_1$, $\lambda_2$, $\lambda_3$, $k$ and $m$. The setting ranges of the corresponding parameters are described in Section~\ref{Settings}. As an example, Figure~\ref{fig:8} (a, b, c) show the sensitivity of the $\lambda_1$, $\lambda_2$ and $\lambda_3$ on the FLICKR25K dataset with 64 bits. Figure~\ref{fig:8} (a) and (b) show that the mAPs are over 0.71 within a large wide ranges w.r.t $\lambda_1$ and $\lambda_2$. In addition, the mAPs fluctuated within a certain range indicate the terms of $\sum_{j=1}^{m}\|\boldsymbol{\rm P}_{j}\|_{2,1}^{2}$ and $\sum_{i,j = 1}^{m}\boldsymbol{\rm S}_{ij}\|\boldsymbol{\rm P}_{i}-\boldsymbol{\rm P}_{j}\|_F$ can influence the performance of the JPSH. We set $\lambda_1$ and $\lambda_2$ as 10 for the FLICKR25K dataset because of the relatively higher and more stable performance. From Figure~\ref{fig:8} (c), mAPs exceed 0.69 in the range of $\{0, 1, \dots, 10^5\}$, and when $\lambda_3$ equals to $10^5$, the mAP surpasses 0.72. Therefore, we set the best $\lambda_3=10^5$ for the FLICKR25K dataset. In addition, Figure~\ref{fig:8} (d) shows mAPs versus $k$ and $m$ on the FLICKR25K dataset with 64 bits, suggesting that the JPSH is not sensitive to $k$ and $m$, either. The sensitivity analysis of $\lambda_1$, $\lambda_2$, $\lambda_3$, $k$ and $m$ on MNIST, CIFAR-10 and NUS-WIDE datasets is similar with that on the FLICKR25K dataset.

\subsection{Convergence Analysis}
Figure~\ref{fig:8} (e) and (f) show the convergence curves of our JPSH method on CIFAR-10 and FLICKR25K datasets with 32 bits. The objective function of the JPSH is well converging, and quickly converges to an optimal solution within 20 iterations in all datasets. From Figure \ref{fig:8} (e), we witness that the objective loss function values settle after 4 iterations. Thus, we set the time iteration of our JPSH for CIFAR-10 dataset as $T=4$. Analogously, based on Figure \ref{fig:8} (f), we set $T=10$ for FLICKR25K dataset. For MNIST and NUS-WIDE datasets, we empirically set iteration times to be $T=10$ and $T=5$, respectively.

\subsection{Time Complexity}
The overall training time complexity of the JPSH is mainly composed by three parts: (1) the building of matrix $\boldsymbol{\rm S}$, which the computational complexity is $O\left(mm\right)$; (2) the time-consuming of computing matrix $\boldsymbol{\rm A}$ is $O\left(nm\right)$; (3) For each iteration, the computational complexity of Eq.~\eqref{eq14} is $O\left(m^3d^3+m^3d^2+m^2dl+mdl^2\right)$; of Eq.~\eqref{eq16} is $O\left(d^3+nd^2+mnd+mdl+dl^2\right)$; of performing SVD on rotation matrices $\boldsymbol{\rm R}$ and $\boldsymbol{\rm V}$ both are $O\left(l^3\right)$; and of Eq.~\eqref{eq23} is $O\left(mdl^2+m^2dl+dl^2+ndl+mnl\right)$. Generally, due to the length of binary codes $l$ and anchor points $m$ are usually smaller, the total computational complexity of our JPSH is $O\left(Td^3+Tnd^2\right)$. It is linear with the number of training samples. Note that the whole computational process is mainly based on matrix multiplications, which can be computed in parallel for accelerations. It is reasonably efficient on the binary representation learning. To quantitatively compare the computational complexity, we evaluate the training time-consumation of OEH, RSSH and JPSH on CIFAR-10 dataset in terms of 32, 64 and 96 bits. Table \ref{tab:3} shows the time complexity and running times of OEH, RSSH and JPSH. Obviously, we could find that JPSH needs much less time than the multi-stage hashing method RSSH while little higher than OEH. It is noted that the JPSH yields higher mAP values than OEH, which are shown in Table~\ref{tab:2}. Therefore, the time cost of our proposed JPSH model is acceptable.

\begin{table}[h]
	\centering
	\caption{Time complexity and running times of OEH, RSSH and JPSH on 32, 64 and 96 bits. Note that the best results are in bold and the second-best results are underlined.}
	\label{tab:3}
	\begin{tabular}{|c|c|c|c|c|c|c|}
		\hline
		\multirow{2}{*}{Methods} &\multirow{2}{*}{Time Complexity} &\multicolumn{3}{c|}{Running Times (seconds)} \\
		\cline{3-5}
		& &32 bits &64 bits &96 bits \\
		\hline
		OEH\cite{liu2016towards}    &$\textgreater O\left(m^2l^3k\right)$ &\textbf{13.3385} &\textbf{21.9931} &\textbf{43.9122}  \\
		RSSH\cite{tian2020unsupervised}   &$\textgreater O\left(Tnm^2+Tnml\right)$ &75.3624 &80.4510 &86.0708  \\
		JPSH   &$O\left(Td^3+Tnd^2\right)$ &\underline{45.0945} &\underline{53.9658} &\underline{62.6788}  \\
		\hline	
	\end{tabular}
\end{table}

\section{Conclusion}
\label{Conclusion}
In this paper, we proposed an effective unsupervised hashing method for binary representation learning, namely Jointly Personalized Sparse Hashing (JPSH), which jointly preserves semantic and pairwise similarities in the Hamming space. Firstly, we developed a novel Personalized Sparse Hashing (PSH) module that adaptively maps different clusters to corresponding personalized subspaces to maintain the identical semantics. Then, we jointed PSH and the manifold-based model JSH to construct JPSH, which can preserve semantic and pairwise similarities in a seamless formulation. Finally, an alternating optimization method was adopted to iteratively solve one variable by fixing the others. Extensive experiments on four benchmark image datasets had been conducted. The results verified that our proposed JPSH outperforms the other state-of-the-art unsupervised hashing methods in the similarity search task.

\begin{acks}
This work was supported in part by  the Guanxi Natural Science Foundation under grants (2019GXNSFFA245014, 2020GXNSFBA238014, 2020GXNSFAA297061), the National Natural Science Foundation of China under grants (62172120, 62002082, 62071174), the Natural Science Foundation of Hunan Province under Grant (2020JJ3014), and the Guangxi Key Research and Development Project (AB21220037).
\end{acks}

\bibliographystyle{ACM-Reference-Format}
\bibliography{ref}

%
%
%
%
%
%
%
%

\end{document}